\documentclass[letterpaper]{article} 
\usepackage{aaai2026}  
\usepackage{times}  
\usepackage{helvet}  
\usepackage{courier}  
\usepackage[hyphens]{url}  
\usepackage{graphicx} 
\usepackage{natbib}  
\usepackage{caption} 
\usepackage{algorithm}
\usepackage{algorithmic}

\usepackage{amsmath}
\usepackage{amssymb}
\usepackage{amsthm}

\usepackage{xcolor}

\newcommand{\first}[1]{\textbf{#1}}
\newcommand{\second}[1]{\underline{#1}}

\usepackage{bbding} 

\usepackage{pifont}

\usepackage[export]{adjustbox}
\usepackage{multirow}
\usepackage{booktabs}
\usepackage{subfig}

\usepackage{pdfpages}

\usepackage{newfloat}
\usepackage{listings}
\DeclareCaptionStyle{ruled}{labelfont=normalfont,labelsep=colon,strut=off} 
\floatstyle{ruled}
\newfloat{listing}{tb}{lst}{}
\floatname{listing}{Listing}
\makeatletter
\def\@fnsymbol#1{\ensuremath{%
  \ifcase#1%
    \or \dagger
    \or *
    \or \ddagger%
    \or \mathsection%
    \or \mathparagraph%
    \or \|%
    \or **%
    \or \dagger\dagger%
    \or \ddagger\ddagger%
  \else\@ctrerr\fi}}
\makeatother

\title{RcAE: Recursive Reconstruction Framework for Unsupervised Industrial Anomaly Detection}
\author{
    Rongcheng Wu\textsuperscript{\rm 2, \rm 6}\equalcontrib,
    Hao Zhu\textsuperscript{\rm 3}\equalcontrib, 
    Shiying Zhang\textsuperscript{\rm 1}\equalcontrib, 
    Mingzhe Wang\textsuperscript{\rm 1}\equalcontrib\thanks{Corresponding author: Mingzhe Wang (E-mail address: wangmingzhe@xidian.edu.cn).}, 
    Zhidong Li\textsuperscript{\rm 2}, 
    Hui Li\textsuperscript{\rm 1},\\
    Jianlong Zhou\textsuperscript{\rm 2}, 
    Jiangtao Cui\textsuperscript{\rm 1}, 
    Fang Chen\textsuperscript{\rm 2}, 
    Pingyang Sun\textsuperscript{\rm 4}, 
    Qiyu Liao\textsuperscript{\rm 3}, 
    Ye Lin\textsuperscript{\rm 2, \rm 5, \rm 6}
}
\affiliations{
    \textsuperscript{\rm 1}School of Computer Science and Technology, Xidian University\\
    \textsuperscript{\rm 2}The Data Science Institute, University of Technology Sydney\\
    \textsuperscript{\rm 3}Data61, CSIRO\\
    \textsuperscript{\rm 4}School of Photovoltaic and Renewable Energy Engineering, University of New South Wales\\
    \textsuperscript{\rm 5}Department of Computing, The Hong Kong Polytechnic University\\
    \textsuperscript{\rm 6}Molly Wardaguga Institute for First Nations Birth Rights, Faculty of Health, Charles Darwin University
}

\usepackage{bibentry}

\begin{document}

\maketitle

\begin{abstract}
Unsupervised industrial anomaly detection requires accurately identifying defects without labeled data. Traditional autoencoder-based methods often struggle with incomplete anomaly suppression and loss of fine details, as their single-pass decoding fails to effectively handle anomalies with varying severity and scale.
We propose a recursive architecture for autoencoder (RcAE), which performs reconstruction iteratively to progressively suppress anomalies while refining normal structures. Unlike traditional single-pass models, this recursive design naturally produces a sequence of reconstructions, progressively exposing suppressed abnormal patterns. To leverage this reconstruction dynamics, we introduce a Cross Recursion Detection (CRD) module that tracks inconsistencies across recursion steps, enhancing detection of both subtle and large-scale anomalies. Additionally, we incorporate a Detail Preservation Network (DPN) to recover high-frequency textures typically lost during reconstruction.
Extensive experiments demonstrate that our method significantly outperforms existing non-diffusion methods, and achieves performance on par with recent diffusion models with only 10\% of their parameters and offering substantially faster inference. These results highlight the practicality and efficiency of our approach for real-world applications.
\end{abstract}


\section{Introduction}

\label{sec:intro}

Anomaly detection is a fundamental task in computer vision with broad applications, such as manufacturing quality control~\cite{MVTEC} and surveillance~\cite{chandola2009anomaly}. In industry, accurate defect detection is critical for ensuring product quality and reducing cost. However, a major challenge arises from severe data imbalance: normal samples are abundant, while anomalies are rare, diverse, and difficult to annotate~\cite{ruff2021unifying}.

\begin{figure}[!t]
	\centering		
    \includegraphics[width=0.87\linewidth]{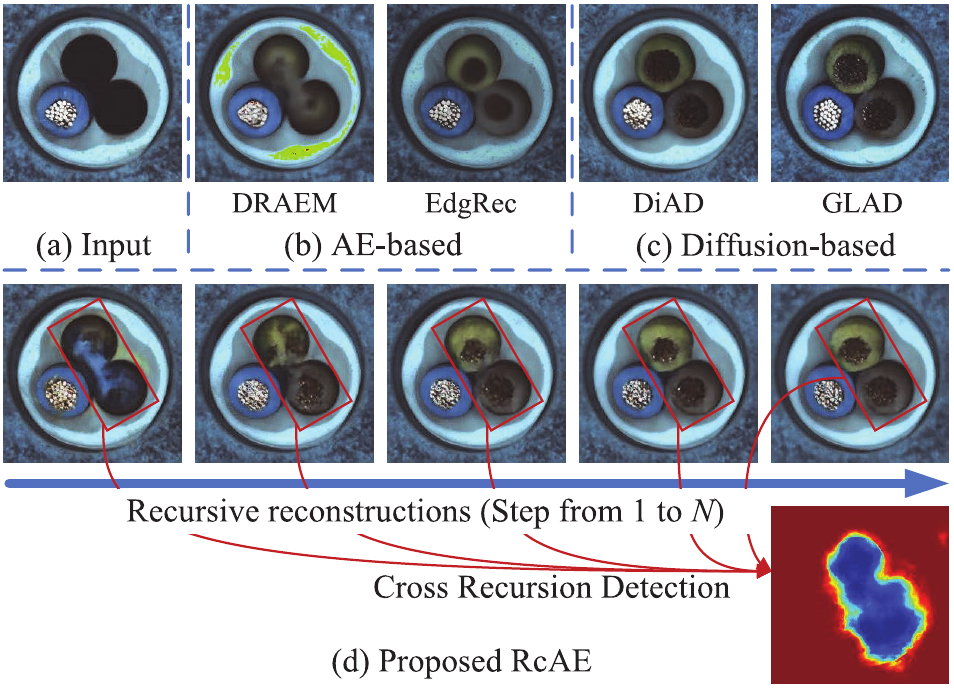}
    \caption{Reconstruction comparison of AE and diffusion-based methods. Our recursive design incrementally suppresses anomalies for high-fidelity reconstruction, then leverages cross-recursion dynamics for accurate detection.}
	\label{F_S1_ours}
\end{figure}

To address this, unsupervised and semi-supervised methods have gained popularity, typically learning the distribution of normal data and detecting deviations at test time as anomalies~\cite{review_2}. Among them, reconstruction-based methods, especially autoencoder (AE) variants, stand out for their simplicity and intuitive design: models trained on normal data are expected to poorly reconstruct unseen anomalies~\cite{reconstruction_1, MemAE}.
However, traditional AE-based methods face several limitations in industrial anomaly detection:
(1) Overfitting to limited and homogeneous normal data~\cite{kingma2019introduction};
(2) Expressive latent spaces may reconstruct anomalies~\cite{bergmann2018improving};
(3) Single-pass decoding that oversmooths fine details~\cite{liu2020towards, reconstruction_1};
(4) Fixed-scale architectures struggle with varying size and severity anomalies~\cite{DREAM, ristea2022self}.

These limitations lead to poor reconstruction quality in traditional AEs. To address this, recent works have introduced GANs~\cite{OCR_GAN}, transformers~\cite{UNIAD}, diffusion models~\cite{Diad, DiffAD}, or augmentation-heavy strategies~\cite{Cutpaste, DREAM,Realnet}. While effective, these methods typically require high computational cost or intricate pre-processing pipelines, limiting their practicality in industrial settings.

In light of these challenges, we revisit a fundamental question: \emph{Can high-quality anomaly reconstruction be achieved without the overhead of complex or resource-heavy designs?} To this end, we propose a lightweight and effective framework based on simple AEs called \textbf{Recursive Convolutional Autoencoder (RcAE)}. As shown in Fig.~\ref{F_S1_ours}, unlike traditional single-pass AEs, RcAE performs iterative reconstruction, progressively suppressing anomalies and reconstructing normal structures across multiple steps. This recursive formulation enables residual evolution to naturally highlight inconsistent anomalies patterns over recursive steps.
To leverage this reconstruction dynamics, we propose a novel \textbf{Cross Recursion Detection (CRD)} module, which monitors reconstruction inconsistencies of anomalies across recursion steps. CRD captures persistent deviations that reveal both micro-defects and large structural anomalies. Additionally, we introduce a lightweight \textbf{Detail Preservation Network (DPN)} to restore fine textures in normal regions that might otherwise be oversmoothed, further reducing false positives caused by detail loss.



Overall, this paper forms an efficient framework, achieving strong detection performance without significantly high computational overhead, making it practical for real-world deployment. \textbf{The main contributions are:}
\begin{itemize}
\item \textbf{Recursive Reconstruction Framework} that performs anomaly suppression and normal pattern enhancement over multiple iterations, enabling significantly robust reconstructions without high computational burden.
\item \textbf{Cross Recursion Detection:} We introduce a novel module that exploits patterns of constantly changing anomalies across recursive steps. By analyzing persistent inconsistencies, Cross Recursion Detection enables unified detection of both subtle and large-scale anomalies.
\item \textbf{Detail Refinement:} While our recursive architecture provide fine reconstructions, we further design a Detail Preservation Network that selectively restores fine textures in normal regions, preserving structural fidelity.
\item \textbf{Efficiency:} Our method achieves state-of-the-art performance on par with recent diffusion models, while requiring 10$\times$ fewer parameters and offering faster inference, making it suitable for industrial deployment.
\end{itemize}


\section{Related Works}
\label{sec_2}

Unsupervised anomaly detection models are trained by only normal data. Reconstruction-based methods remain prominent due to their intuitive logic: a model that learns to reconstruct normal patterns should struggle to reconstruct unseen anomalies, thus exposing them by reconstruction errors.

\vspace{0.1cm}
\noindent\textbf{Autoencoders: Foundations and Challenges.} A classical realization of this paradigm is the autoencoder (AE) framework, which learns to compress and reconstruct normal data~\cite{MemAE,AE_2,AE_3}. While simple and effective in low-data regimes, traditional AEs exhibit several persistent drawbacks:
(1) They easily overfit on limited and homogeneous normal data, leading to poor generalization and high false negatives~\cite{kingma2019introduction}; 
(2) Expressive latent spaces may reconstruct anomalous regions, diminishing detection contrast and accuracy~\cite{bergmann2018improving}; 
(3) Single-pass decoding often smooths out high-frequency details, causing false positives in normal areas~\cite{liu2020towards, reconstruction_1}; 
(4) Fixed-scale architectures struggle with anomalies of varying size and severity, limiting robustness across diverse defect types~\cite{DREAM, ristea2022self}.
These limitations expose the fragility of single-pass reconstruction mechanisms, especially in high-precision use cases.

\vspace{0.1cm}
\noindent\textbf{Beyond AEs: Better Reconstructions at a Cost.} To address the reconstruction limitations of basic AEs, recent research has explored more expressive models:
\textbf{GAN-based approaches}~\cite{OCR_GAN,GAN_1} introduce adversarial losses to generate visually sharper reconstructions. While enhancing realism, they often suffer from training instability and require large-scale data~\cite{akcay2019ganomaly}, limiting their practicality in sparse-label industrial scenarios.
\textbf{Transformer-based models}~\cite{UNIAD,Transformer_1,Transformer_2} leverage self-attention to capture long-range dependencies. For instance, UniAD~\cite{UNIAD} unifies multiple object categories under a single framework. However, their high memory cost and complex optimization hinder deployment in real-time applications~\cite{xu2021anomaly}.
\textbf{Diffusion-based frameworks}~\cite{Diad,GLAD,DiffAD,DDPM} have recently achieved state-of-the-art reconstruction quality by learning denoising trajectories from noise to clean samples. Their iterative nature enables gradual normalization of anomalies, making them powerful but computationally prohibitive, often requiring dozens to hundreds of denoising steps per image~\cite{LDM}. Moreover, diffusion models may still suffer from semantic inconsistencies and detail loss~\cite{Efficientad, reconstruction_1}.
These reflects a clear trend: improvements in reconstruction typically comes at the cost of inference latency, training complexity, or resource requirements, which is a tradeoff that undermines deployment in resource-constrained industrial settings.

\begin{figure*}[t]
    \centering
    \includegraphics[width=1\linewidth]{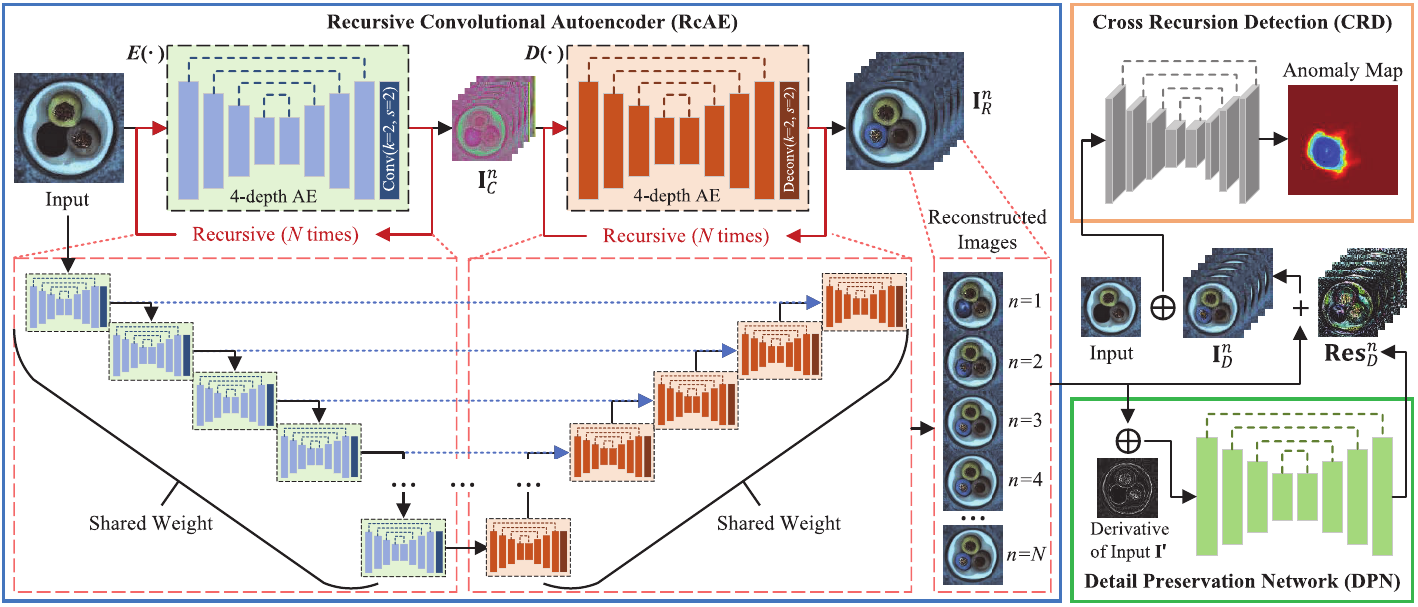}
    \caption{Overview of the proposed anomaly detection framework. 
    RcAE performs iterative reconstruction to progressively suppress anomalies and stabilize normal structures; 
    DPN selectively restores high-frequency textures; 
    CRD leverages reconstruction dynamics across recursion steps to produce robust pixel-level anomaly maps.}    
    \label{fig:complete_pipeline}
\end{figure*}

\vspace{0.1cm}
\noindent\textbf{Limitations and Our Perspective.} 
Despite ongoing progress, reconstruction-based methods still face several persistent challenges: 
(1) Loss of fine details in normal regions, leading to false positives~\cite{denouden2018improving}; 
(2) Incomplete suppression of anomalies, causing false negatives~\cite{cai2024rethinking}; 
(3) Loose boundary modeling of normal distributions, allowing reconstruction of anomalies~\cite{review_1}; 
(4) High computational demands in high-performing models, leading to poor deployment feasibility~\cite{fuvcka2024transfusion}. 
Thus, improving reconstruction quality without incurring high cost is still a challenging issue, and we aim to propose a framework that is both high-performing and computationally efficient, striking a better balance for practical deployment.


\section{Methodology}

As illustrated in Fig.~\ref{fig:complete_pipeline}, the proposed framework is a lightweight yet expressive reconstruction-based pipeline with three key components: 
(1) a \textbf{Recursive Autoencoder (RcAE)} that reformulates reconstruction as a multi-step refinement process, progressively suppressing anomalies while stabilizes normal structures across iterations; 
(2) a \textbf{Detail Preservation Network (DPN)} that restores high-frequency details in normal regions, reducing false positives while preserving structural fidelity; and
(3) a \textbf{Cross Recursion Detection (CRD)} that exploits reconstruction dynamics across recursion steps to highlight persistent inconsistencies, generating robust anomaly detection.


\subsection{Preliminaries}
\label{sec:preliminaries}

Let input space $\mathcal{X} \subset \mathbb{R}^{H \times W \times C}$ contain images with $H \times W$ resolution and $C$ channels. The training set $\mathcal{D}_{\text{train}} = \{\mathbf{x}_1, \mathbf{x}_2, \ldots, \mathbf{x}_S\}$ contains only normal images $\mathbf{x}_i \in \mathcal{X}$. The goal of unsupervised anomaly detection is to learn a mapping $f: \mathcal{X} \rightarrow \mathcal{X}$ such that reconstruction errors $|\mathbf{x}_i - f(\mathbf{x}_i)|$ expose anomalies in unseen test images. During inference, an anomaly score is assigned to each pixel of a test image $\mathbf{x}_{\text{test}} \in \mathcal{X}$ based on deviations from learned normal patterns. 

\noindent\textbf{Standard Autoencoder Framework.}
A traditional autoencoder consists of an encoder $E_{\boldsymbol{\theta}_E}: \mathcal{X} \rightarrow \mathcal{Z}$ and a decoder $D_{\boldsymbol{\theta}_D}: \mathcal{Z} \rightarrow \mathcal{X}$, where $\mathcal{Z} \subset \mathbb{R}^{d}$ is a compact latent space with dimensionality $d \ll H \times W \times C$, $\boldsymbol{\theta}_E$ and $\boldsymbol{\theta}_D$ are the parameters. The network is trained to minimize the reconstruction loss on normal samples:
\begin{equation}\textstyle
\mathcal{L}_{\text{rec}} = \frac{1}{N} \sum_{i=1}^{N} \|\mathbf{x}_i - D_{\boldsymbol{\theta}_D}(E_{\boldsymbol{\theta}_E}(\mathbf{x}_i))\|_p,
\end{equation}
where $\|\cdot\|_p$ denotes the $\ell_p$ norm. The reconstruction of an input image $x$ is denoted as $\hat{x} = D_{\boldsymbol{\theta}_D}(E_{\boldsymbol{\theta}_E}(\mathbf{x}))$. The AE-based methods are simple and efficient, but often fail to fully suppress subtle anomalies and preserve fine details. 

\subsection{Recursive Convolutional Autoencoder}
\label{subsec:recursive_autoencoder}

Traditional deep convolutional autoencoders (ConvAEs) improve reconstruction capacity by stacking $N$ distinct encoder and decoder blocks. Formally, the encoder blocks $\{E_1, E_2, ..., E_N\}$ with parameters $\{\boldsymbol{\theta}_{E_1}, \boldsymbol{\theta}_{E_2}, ..., \boldsymbol{\theta}_{E_N}\}$ and decoder blocks $\{D_1, D_2, ..., D_N\}$ with parameters $\{\boldsymbol{\theta}_{D_1}, \boldsymbol{\theta}_{D_2}, ..., \boldsymbol{\theta}_{D_N}\}$, define a mapping:
\begin{equation}
f_{\text{deep}}(\mathbf{x}) = D_1 \circ D_2 \circ ... \circ D_N \circ E_N \circ ... \circ E_2 \circ E_1(\mathbf{x}),
\end{equation}
where each block has it own parameters. While effective but significantly increases model size and training complexity.

To achieve high-quality reconstruction without inflating the parameter count, we propose a Recursive Convolutional Autoencoder (RcAE) that replaces these stacked layers with a compact, recursive design. Instead of using multiple independent blocks, RcAE reuses two shared autoencoders in an iterative pipeline to perform progressive compression and reconstruction, mimicking the depth and semantic abstraction of a deep ConvAE with far fewer parameters.

\vspace{0.1cm}
\noindent\textbf{Recursive Structure.} As shown in Fig.~\ref{fig:complete_pipeline}, RcAE consists of two phases: a recursive compression phase with encoder $E(\cdot)$ and a recursive reconstruction phase with decoder $D(\cdot)$, both adopt a standard 4-layer ConvAE with skip connections and parameter sharing within each phase.
To simulate the behavior of a deeper AE while preserving the scale-space encoding and decoding, we insert a downsampling convolution layer (kernel size = 2, stride = 2) after $E(\cdot)$, reducing spatial resolution by a factor of 2. Likewise, $D(\cdot)$ ends with a deconvolutional upsampling layer (kernel size = 2, stride = 2) to restore resolution. This allows each recursion operates at progressively coarser semantic levels.

\vspace{0.1cm}
\noindent\textbf{Compression Phase.} 
Given input image $\mathbf{x}$, the compression phase progressively compresses it through $N$ recursions using a shared encoder $E$ with parameters $\boldsymbol{\theta}_E$:
\begin{equation}
\mathbf{I}_C^i = E(\mathbf{I}_C^{i-1}; \boldsymbol{\theta}_E), \quad i \in \{1, 2, ..., N\},
\end{equation}
where $\mathbf{I}_C^0 = \mathbf{x}$. Each recursion step reduces spatial resolution by the shared encoder $E: \mathcal{X}_{i-1} \rightarrow \mathcal{X}_i$ maps from one resolution level to a lower one, where $\mathcal{X}_i \subset \mathbb{R}^{H/2^i \times W/2^i \times C}$, enabling hierarchical abstraction of visual content.

\vspace{0.1cm}
\noindent\textbf{Reconstruction Phase.}
Starting from the compressed representation $\mathbf{I}_C^N$ from $N$-times recursion, the reconstruction phase progressively reconstructs the image through $N$ recursions using a shared decoder $D$ with parameters $\boldsymbol{\theta}_D$:
\begin{equation}
\mathbf{I}_R^j = D(\mathbf{I}_R^{j-1}; \boldsymbol{\theta}_D), \quad j \in \{1, 2, ..., N\},
\end{equation}
where $\mathbf{I}_R^0 = \mathbf{I}_C^N$, and each iteration restores the spatial resolution by the shared decoder $D: \mathcal{X}_{N-j+1} \rightarrow \mathcal{X}_{N-j}$ which maps from one resolution level to a higher one. The final reconstruction $\mathbf{I}_R^N$ represents the full-resolution anomaly-normalized output.

\vspace{0.1cm}
\noindent\textbf{Progressive Reconstruction Benefits.}
Together, these components implement a recursive architecture where deeper iteration compresses and reconstructs the image at a coarser scale, effectively performing reconstruction at multiple semantic levels: early iterations retain low-level details but may still contain residual anomalies, while later iterations better suppress anomalies but may over-smooth fine structures. Such progressive refinement allows RcAE to simultaneously suppress anomalies and preserve normal structures, improving robustness without extra parameters.


\vspace{0.1cm}
\noindent\textbf{Training.}
To encourage generalization across varying recursive depths and adapt to anomalies of different intensities, the number of recursions is randomly sampled from $[1, N]$ during training, which also help avoiding shortcut learning. We supervise both intensity and edge fidelity as follows:
\begin{equation}
\label{eq:rcAE_loss}
\begin{aligned}
\mathcal{L}_\text{rec} = \|\mathbf{I} - \mathbf{I}_R^{N}\|_1+\|\mathbf{I}' - \mathbf{I}'^{N}_R\|_1,
\end{aligned}
\end{equation}
where $\mathbf{I}'$ and $\mathbf{I}'^{N}_R$ denote the first-order gradients of input and the final reconstruction, respectively.

\subsection{Detail Preservation Network}
\label{subsec:detail_preservation}

While RcAE effectively suppresses anomalies through multi-step reconstruction, the recursive nature of the process can also lead to the accumulation of detail loss in normal regions, leading to false positives.

To address this, we propose a Detail Preservation Network (DPN), which selectively restores fine details in normal areas while avoiding the reintroduction of anomalies. As illustrated in the 2nd stage of Fig.~\ref{fig:complete_pipeline}, DPN is a lightweight 4-layer convolutional autoencoder with skip connections, which takes the recursive reconstructions $\{\mathbf{I}_R^n\}$ concatenated with the first-order derivative of the input $\mathbf{I}'$, to predict residual maps $\{\mathbf{Res}^n_D\}$ contains missing details:
\begin{equation}
\label{eq:detail_dpn}
\begin{aligned}
\mathbf{Res}^n_D = f_{\text{DPN}}((\mathbf{I}_R^n \oplus \mathbf{I}'); \boldsymbol{\theta}_{\text{DPN}}),\quad\mathbf{I}^n_D = \mathbf{I}_R^n + \mathbf{Res}^n_D,
\end{aligned}
\end{equation}
where $\oplus$ denotes channel-wise concatenation, and $\boldsymbol{\theta}_{\text{DPN}}$ is the parameters. The output $\mathbf{I}^n_D$ represents the detail-enhanced reconstruction for downstream anomaly scoring.

DPN is supervised by a dual-term $\ell_1$ loss to recover both intensity and edge details:
\begin{equation}
\label{eq:detail_loss}
\begin{aligned}
\mathcal{L}_{\text{DPN}}=\|(\mathbf{Res}^n_D + \mathbf{I}_R^n) - \mathbf{I}\|_1 + \|(\mathbf{Res}^n_D + \mathbf{I}^n_R)' - \mathbf{I}' \|_1,
\end{aligned}
\end{equation}
where $\mathbf{I}'$ and $(\mathbf{Res}^n_D + \mathbf{I}^n_R)'$ denote the gradient maps of $\mathbf{I}$ and $\mathbf{I}^n_D$, respectively.

Importantly, RcAE is frozen during DPN training, and only clean normal samples are used. This forces the network to focus on learning residuals caused by recursive detail degradation rather than anomaly-related deviations. In inference, since anomalies generate unfamiliar residuals outside the learned distribution, DPN naturally fails to restore them, preserving anomaly suppression. This selective recovery mechanism effectively enhances fine details in normal regions while reducing false positives, thereby improving the reliability of pixel-wise anomaly detection.

\begin{table*}[!t]
	\centering
	\begin{adjustbox}{max width=1\textwidth}
\begin{tabular}{@{}ccc|ccccc|ccc|c|c@{}}
\toprule
\multicolumn{3}{c|}{\multirow{2}{*}{Category}}                                                                                                        & \multicolumn{5}{c|}{Non-Diffusion Method}                                                                                                                 & \multicolumn{3}{c|}{Diffusion-based}                                  & Flow-based                         & Diffusion+DINO                      \\ \cmidrule(l){4-13} 
\multicolumn{3}{c|}{}                                                                                                                                 & DRAEM                      & PatchCore                   & RD4AD                     & EfficientAD                 & \textbf{Ours}                        & D3AD               & DiAD               & DiffAD                      & MSFlow                             & GLAD                                \\ \midrule
\multicolumn{3}{c|}{From Scratch?}                                                                                                                    & \ding{51}                  & \ding{55}                   & \ding{55}                 & \ding{55}                   & \ding{51}                            & \ding{55}          & \ding{55}          & \ding{51}                   & \ding{55}                          & \ding{55}                           \\ \midrule
\multicolumn{1}{c|}{\multirow{15}{*}{\rotatebox{90}{MVTec AD Dataset}}} & \multicolumn{1}{c|}{\multirow{10}{*}{\rotatebox{90}{Objects}}} & Bottle     & 99.2/\first{99.1}          & \first{100}/98.6            & \first{100}/\second{99.0} & \second{99.9}/98.7          & \first{100}/\first{99.1}             & \first{100}/98.6   & 99.7/98.4          & \first{100}/98.8            & \first{100}/\second{99.0}          & \first{100}/98.9                    \\
\multicolumn{1}{c|}{}                                                   & \multicolumn{1}{c|}{}                                          & Cable      & 91.8/94.7                  & \second{99.5}/98.4          & 95.0/\first{99.4}         & 95.2/\second{98.8}          & 97.4/97.1                            & 97.8/93.3          & 94.8/96.8          & 94.6/96.8                   & \second{99.5}/98.5                 & \first{99.9}/98.1                   \\
\multicolumn{1}{c|}{}                                                   & \multicolumn{1}{c|}{}                                          & Capsule    & 98.5/94.3                  & 98.1/\second{98.8}          & 96.3/97.3                 & 97.9/\first{99.2}           & 94.4/97.9                            & 96.6/97.9          & 89.0/97.1          & 97.5/98.2                   & \second{99.2}/98.1                 & \first{99.5}/98.5                   \\
\multicolumn{1}{c|}{}                                                   & \multicolumn{1}{c|}{}                                          & Hazelnut   & \first{100}/\first{99.7}   & \first{100}/98.7            & \second{99.9}/98.2        & 99.4/98.8                   & \first{100}/\second{99.5}            & 98.0/98.8          & 99.5/98.3          & \first{100}/99.4            & \first{100}/98.7                   & \first{100}/\second{99.5}           \\
\multicolumn{1}{c|}{}                                                   & \multicolumn{1}{c|}{}                                          & Metal Nut  & 98.7/\second{99.5}         & \first{100}/98.4            & \first{100}/\first{99.6}  & 99.6/98.5                   & \second{99.8}/98.8                   & 98.9/96.1          & 99.1/97.3          & \first{100}/99.4            & \first{100}/99.3                   & \first{100}/98.8                    \\
\multicolumn{1}{c|}{}                                                   & \multicolumn{1}{c|}{}                                          & Pill       & 98.9/97.6                  & 96.6/97.4                   & 96.6/95.7                 & 98.6/98.7                   & 98.4/\first{98.9}                    & \second{99.2}/98.2 & 95.7/95.7          & 97.7/97.7                   & \first{99.6}/\second{98.8}         & 98.1/97.9                           \\
\multicolumn{1}{c|}{}                                                   & \multicolumn{1}{c|}{}                                          & Screw      & 93.9/97.6                  & 97.0/\first{99.1}           & 97.0/\first{99.1}         & 97.0/98.7                   & 95.8/98.7                            & 83.9/\second{99.0} & 90.7/97.9          & \second{97.2}/\second{99.0} & \first{97.8}/\first{99.1}          & 96.9/\first{99.1}                   \\
\multicolumn{1}{c|}{}                                                   & \multicolumn{1}{c|}{}                                          & Toothbrush & \first{100}/98.1           & \first{100}/98.7            & 99.5/93.0                 & \first{100}/97.7            & \first{100}/\first{99.4}             & \first{100}/99.0   & \second{99.7}/99.0 & \first{100}/\second{99.2}   & \first{100}/98.5                   & \first{100}/\first{99.4}            \\
\multicolumn{1}{c|}{}                                                   & \multicolumn{1}{c|}{}                                          & Transistor & 93.1/90.9                  & \first{100}/96.3            & 96.7/95.4                 & \second{99.9}/\second{97.2} & 98.6/96.6                            & 96.8/95.6          & 99.8/95.1          & 96.1/93.7                   & \first{100}/\first{98.3}           & 98.3/96.2                           \\
\multicolumn{1}{c|}{}                                                   & \multicolumn{1}{c|}{}                                          & Zipper     & \first{100}/98.8           & 99.4/98.8                   & 98.5/98.2                 & \second{99.7}/96.3          & \first{100}/\first{99.6}             & 98.2/98.3          & 95.1/96.2          & \first{100}/99.0            & \first{100}/\second{99.2}          & 98.5/97.9                           \\ \cmidrule(l){2-13} 
\multicolumn{1}{c|}{}                                                   & \multicolumn{1}{c|}{\multirow{5}{*}{\rotatebox{90}{Texture}}}  & Carpet     & 97.0/95.5                  & 98.7/99.0                   & 98.9/98.8                 & 99.3/96.3                   & \first{100}/\first{99.6}             & 94.2/97.6          & \second{99.4}/98.6 & 98.3/98.1                   & \first{100}/\second{99.4}          & 99.0/98.5                           \\
\multicolumn{1}{c|}{}                                                   & \multicolumn{1}{c|}{}                                          & Grid       & \second{99.9}/\first{99.7} & 98.2/98.7                   & \first{100}/97.0          & 99.9/94.1                   & \first{100}/\first{99.7}             & \first{100}/99.2   & 98.5/96.6          & \first{100}/\first{99.7}    & 99.8/99.4                          & \first{100}/\second{99.6}           \\
\multicolumn{1}{c|}{}                                                   & \multicolumn{1}{c|}{}                                          & Leather    & \first{100}/98.6           & \first{100}/99.3            & \first{100}/98.6          & \first{100}/97.7            & \first{100}/\first{99.8}             & 98.5/99.4          & \second{99.8}/98.8 & \first{100}/99.1            & \first{100}/\second{99.7}          & \first{100}/\first{99.8}            \\
\multicolumn{1}{c|}{}                                                   & \multicolumn{1}{c|}{}                                          & Tile       & 99.6/\second{99.2}         & 98.7/95.6                   & 99.3/98.9                 & \second{99.9}/91.5          & 99.2/97.8                            & 95.5/94.7          & 96.8/92.4          & \first{100}/\first{99.4}    & \first{100}/98.2                   & \first{100}/98.7                    \\
\multicolumn{1}{c|}{}                                                   & \multicolumn{1}{c|}{}                                          & Wood       & 99.1/96.4                  & 99.2/95.0                   & 99.2/\first{99.3}         & 99.5/90.9                   & \first{100}/97.9                     & \second{99.7}/95.9 & \second{99.7}/93.3 & \first{100}/96.7            & \first{100}/97.1                   & 99.4/\second{98.4}                  \\ \cmidrule(l){2-13} 
\multicolumn{1}{c|}{}                                                   & \multicolumn{2}{c|}{\textbf{Avg.}}                                          & {98.0/97.3}         & {99.1/98.1}          & {98.5/97.8}        & {99.1/96.9}          & {98.9/\second{98.7}}          & {97.2/97.4} & {97.2/96.8} & {98.7/98.3}          & {\first{99.7}/\first{98.8}} & {\second{99.3}/98.6}         \\ \midrule
\multicolumn{1}{c|}{\multirow{13}{*}{\rotatebox{90}{VisA Dataset}}}     & \multicolumn{2}{c|}{Candle}                                                 & 89.6/91.0                  & \second{98.7}/\second{99.2} & 94.3/98.7                 & 98.4/99.1                   & \first{99.9}/\first{99.3}            & 95.6/-             & 92.8/97.3          & 90.4/-                      & 97.7/98.3                          & \first{99.9}/94.8                   \\
\multicolumn{1}{c|}{}                                                   & \multicolumn{2}{c|}{Capsules}                                               & 89.2/99.0                  & 68.8/96.5                   & 90.8/\second{99.4}        & 93.5/98.2                   & \second{98.7}/\first{99.6}           & 88.5/-             & 58.2/97.3          & 87.6/-                      & 98.0/96.2                          & \first{99.1}/\first{99.6}           \\
\multicolumn{1}{c|}{}                                                   & \multicolumn{2}{c|}{Cashew}                                                 & 88.3/85.0                  & \second{97.7}/\first{99.2}  & 97.4/94.1                 & 97.2/\first{99.2}           & 96.9/96.2                            & 94.2/-             & 91.5/90.9          & 81.4/-                      & 94.9/\second{98.7}                 & \first{98.4}/97.0                   \\
\multicolumn{1}{c|}{}                                                   & \multicolumn{2}{c|}{Chewinggum}                                             & 96.4/97.7                  & 99.1/98.9                   & 98.4/97.4                 & \first{99.9}/99.2           & \second{99.8}/\second{99.4}          & 99.7/-             & 99.1/94.7          & 94.0/-                      & 93.6/\first{99.7}                  & 99.6/99.1                           \\
\multicolumn{1}{c|}{}                                                   & \multicolumn{2}{c|}{Fryum}                                                  & 94.7/82.5                  & 91.6/95.9                   & 96.2/96.7                 & 96.5/96.5                   & \first{99.9}/\second{97.6}           & 96.5/-             & 89.8/\second{97.6} & 87.1/-                      & 88.2/\first{99.6}                  & \second{99.4}/96.9                  \\
\multicolumn{1}{c|}{}                                                   & \multicolumn{2}{c|}{Macaroni1}                                              & 93.9/99.4                  & 90.1/98.5                   & 98.6/99.6                 & 99.4/\first{99.9}           & \second{99.7}/99.3                   & 94.3/-             & 85.7/94.1          & 87.6/-                      & 97.6/97.6                          & \first{99.9}/\second{99.8}          \\
\multicolumn{1}{c|}{}                                                   & \multicolumn{2}{c|}{Macaroni2}                                              & 88.3/\second{99.7}         & 63.4/93.5                   & 89.5/99.2                 & 96.7/99.8                   & 97.1/99.5                            & 92.5/-             & 62.5/93.6          & 90.7/-                      & \second{98.0}/89.5                 & \first{98.9}/\first{99.8}           \\
\multicolumn{1}{c|}{}                                                   & \multicolumn{2}{c|}{Pcb1}                                                   & 84.7/98.4                  & 96.0/\first{99.8}           & 97.1/\second{99.7}        & 98.5/\first{99.8}           & \first{99.7}/\second{99.7}           & 97.7/-             & 88.1/98.7          & 75.0/-                      & 96.0/98.9                          & \second{99.6}/99.6                  \\
\multicolumn{1}{c|}{}                                                   & \multicolumn{2}{c|}{Pcb2}                                                   & 96.2/94.0                  & 95.1/98.4                   & 97.0/\second{98.6}        & \second{99.5}/\first{99.3}  & \second{99.5}/98.1                   & 98.3/-             & 91.4/95.2          & 94.6/-                      & 93.5/97.8                          & \first{100}/\second{98.6}           \\
\multicolumn{1}{c|}{}                                                   & \multicolumn{2}{c|}{Pcb3}                                                   & 97.4/94.3                  & 93.0/98.9                   & 96.4/\second{99.2}        & 98.9/\first{99.4}           & \second{99.5}/98.3                   & 97.4/-             & 86.2/96.7          & 94.7/-                      & 94.4/98.9                          & \first{99.9}/98.9                   \\
\multicolumn{1}{c|}{}                                                   & \multicolumn{2}{c|}{Pcb4}                                                   & 98.9/97.6                  & 99.5/98.3                   & \first{99.9}/97.7         & 98.9/\second{99.1}          & \first{99.9}/98.7                    & \second{99.8}/-    & 99.6/97.0          & 97.7/-                      & 93.0/\first{99.5}                  & \first{99.9}/\first{99.5}           \\
\multicolumn{1}{c|}{}                                                   & \multicolumn{2}{c|}{Pipe fryum}                                             & 94.7/65.8                  & 99.0/\second{99.3}          & 94.6/98.7                 & \second{99.7}/\second{99.3} & \first{99.9}/97.5                    & 96.9/-             & 96.2/\first{99.4}  & 92.7/-                      & 97.0/98.9                          & 98.9/\first{99.4}                   \\ \cmidrule(l){2-13} 
\multicolumn{1}{c|}{}                                                   & \multicolumn{2}{c|}{\textbf{Avg.}}                                          & {92.4/92.0}         & {91.0/98.1}          & {95.8/98.3}        & {98.1/\first{99.1}}  & {\second{99.2}/\second{98.6}} & {96.0/97.9} & {86.8/96.0} & {89.5/-}             & {95.2/97.8}                 & {\first{99.5}/\second{98.6}} \\ \midrule
\multicolumn{3}{c|}{\textbf{Avg. All}}                                                                                                                & {95.6/94.9}         & {95.4/98.0}          & {97.3/98.0}        & {98.6/97.8}          & {\second{99.0}/\first{98.7}}  & {96.6/97.4} & {92.5/96.4} & {94.6/-}            & {97.7/98.3}                 & {\first{99.4}/\second{98.6}} \\ \bottomrule
\end{tabular}
	\end{adjustbox}
    \caption{Comparison of anomaly detection and localization performance on MVTec AD and VisA. Each entry reports I-AUROC / P-AUROC (\%). \first{Bold} and \second{underlined} numbers indicate the \first{best} and \second{second-best} results, respectively.}
    \label{T_comparsion_exp}
\end{table*}

\subsection{Cross Recursion Detection}
\label{subsec:anomaly_detection_pipeline}

Our recursive design naturally produces a sequence of reconstructions, and the differences between steps reflect region-wise stability, i.e., normal regions stabilize quickly, while anomalous regions fluctuate due to reconstruction difficulty. Early iterations may retain residual defects but preserve fine details, whereas later iterations suppress anomalies better but may lose subtle textures. To leverage this reconstruction dynamics, we introduce the Cross Recursion Detection (CRD) for robust anomaly localization.

As shown in the 3rd stage of Fig.~\ref{fig:complete_pipeline}, CRD is a 4-depth 3D ConvAE with skip connections that jointly models spatial features and reconstruction dynamics across recursion steps. It takes the concatenation of the input $\mathbf{I}$ and the detail-enhanced reconstructions ${\mathbf{I}_D^n}$ to predict anomaly map $\mathbf{M}_A$:
\begin{equation}
\label{eq:detail_MSD}
\begin{aligned}
\mathbf{M}_A = f_{\text{CRD}}((\mathbf{I}_D^n \oplus \mathbf{I}); \boldsymbol{\theta}_{\text{CRD}}),\quad n \in \{1, 2, ..., N\},
\end{aligned}
\end{equation}
where $\boldsymbol{\theta}_{\text{CRD}}$ are the learnable parameters. 3D convolutions allow CRD to extract cross-recursion temporal patterns, highlighting regions that remain unstable across iterations.


During training, both RcAE and DPN are frozen. We use only normal images, and generate pseudo anomaly masks $\mathbf{M}_P$ via simple augmentations (e.g., color patches, random lines, copy-paste). CRD is optimized using a dual-term $\ell_2$ loss for spatial and edge consistency:
\begin{equation}
\label{eq:MSD_loss}
\begin{aligned}
\mathcal{L}_{\text{CRD}}=\|\mathbf{M}_A - \mathbf{M}_P\|_2 + \|\mathbf{M}'_A - \mathbf{M}'_P\|_2,
\end{aligned}
\end{equation}
where $\mathbf{M}'$ denotes the gradient map. At test time, CRD outputs the final pixel-wise anomaly map $\mathbf{M}_A$. For image-level anomaly detection, we follow standard practice of averaging the top-$k$ pixel scores.

In contrast to prior methods that rely solely on a single reconstruction, our CRD module fully exploits cross-recursion dynamics of RcAE, offering reliable detection of anomalies at multiple scales and varying intensities.

\subsection{Training Strategy}
\label{subsec:training}

Our framework is trained in three independent stages on normal data to ensure stability and modular effectiveness. Note that all components are trained from scratch:

\begin{itemize}
    \item \noindent\textbf{Stage 1:} Train RcAE with $\mathcal{L}_\text{rec}$. To prevent shortcut learning and overfitting to a fixed recursion depth, the recursion depth is randomly selected from $[1, N]$ per batch.

\item \noindent\textbf{Stage 2:} Freeze RcAE, train DPN with loss $\mathcal{L}_{\text{DPN}}$ to  restore high-frequency details via $\mathbf{I}_D^n = \mathbf{I}_R^n + \mathbf{Res}_D^n$.

\item \noindent\textbf{Stage 3:} Freeze RcAE and DPN, train CRD with $\mathcal{L}_{\text{CRD}}$ using $\mathbf{I}$ and $\{\mathbf{I}_D^n\}$ to predict anomaly map $\mathbf{M}_A$. Pseudo masks $\mathbf{M}_P$ are generated via lightweight augmentations.

\end{itemize}

\noindent\textbf{Augmentation:}  
We adopt simple perturbations in Stages 1 and 3, including:
(1) Random color blocks, (2) Copy \& paste patches, (3) Random lines (1–4) of length 50–150 forming crack-like structures. These are applied to random blocks of size 32, 64, or 128 with varying coverage (0–100\%).


\section{Experiments}

\label{sec_4}

\begin{figure*}[!t]
	\centering
    \captionsetup[subfigure]{justification=centering,singlelinecheck=false}
	\subfloat[Inputs]{
        \begin{tabular}{@{}p{0.93in}@{}}\centering
			\includegraphics[width=0.79in]{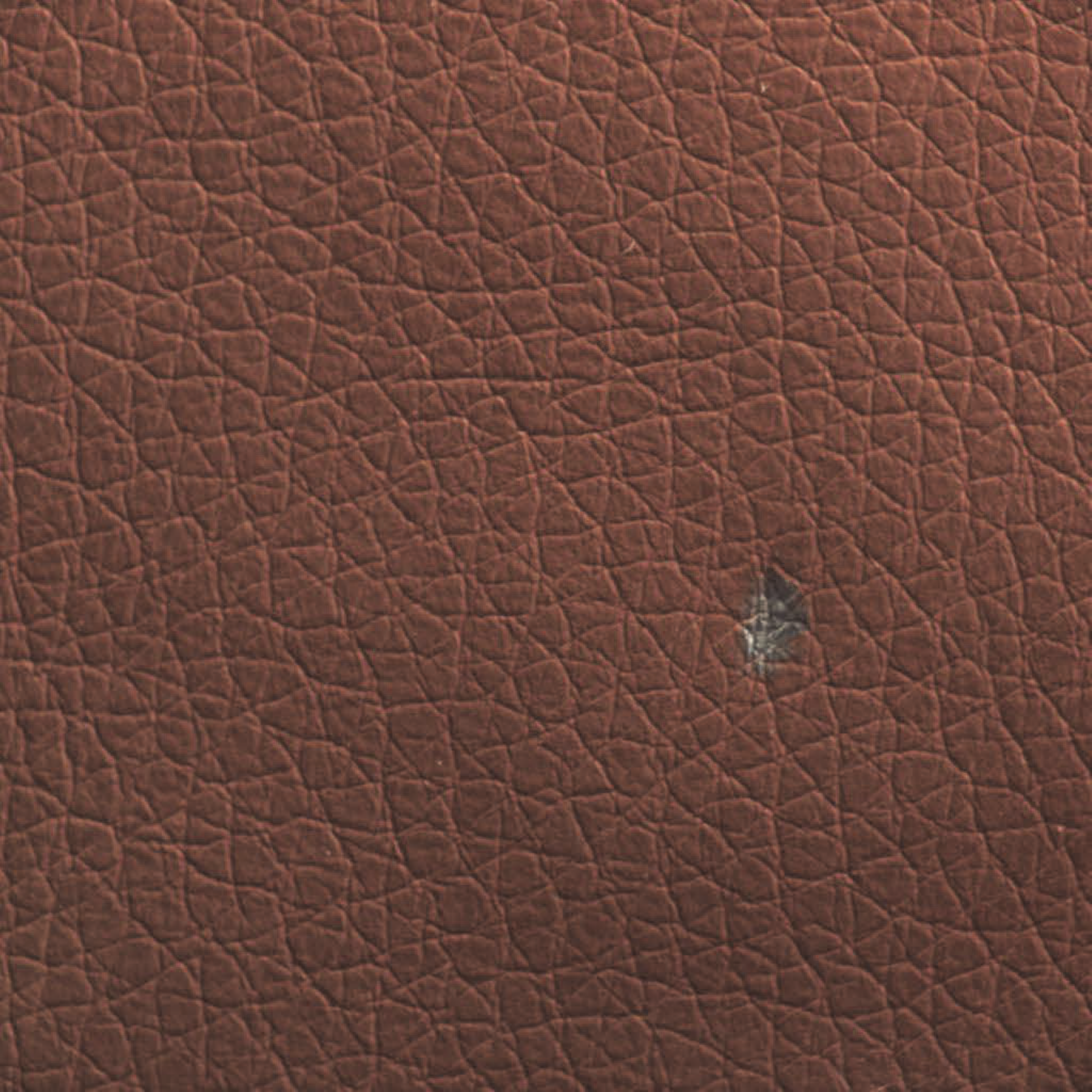}\\[1.5pt]
            \includegraphics[width=0.79in]{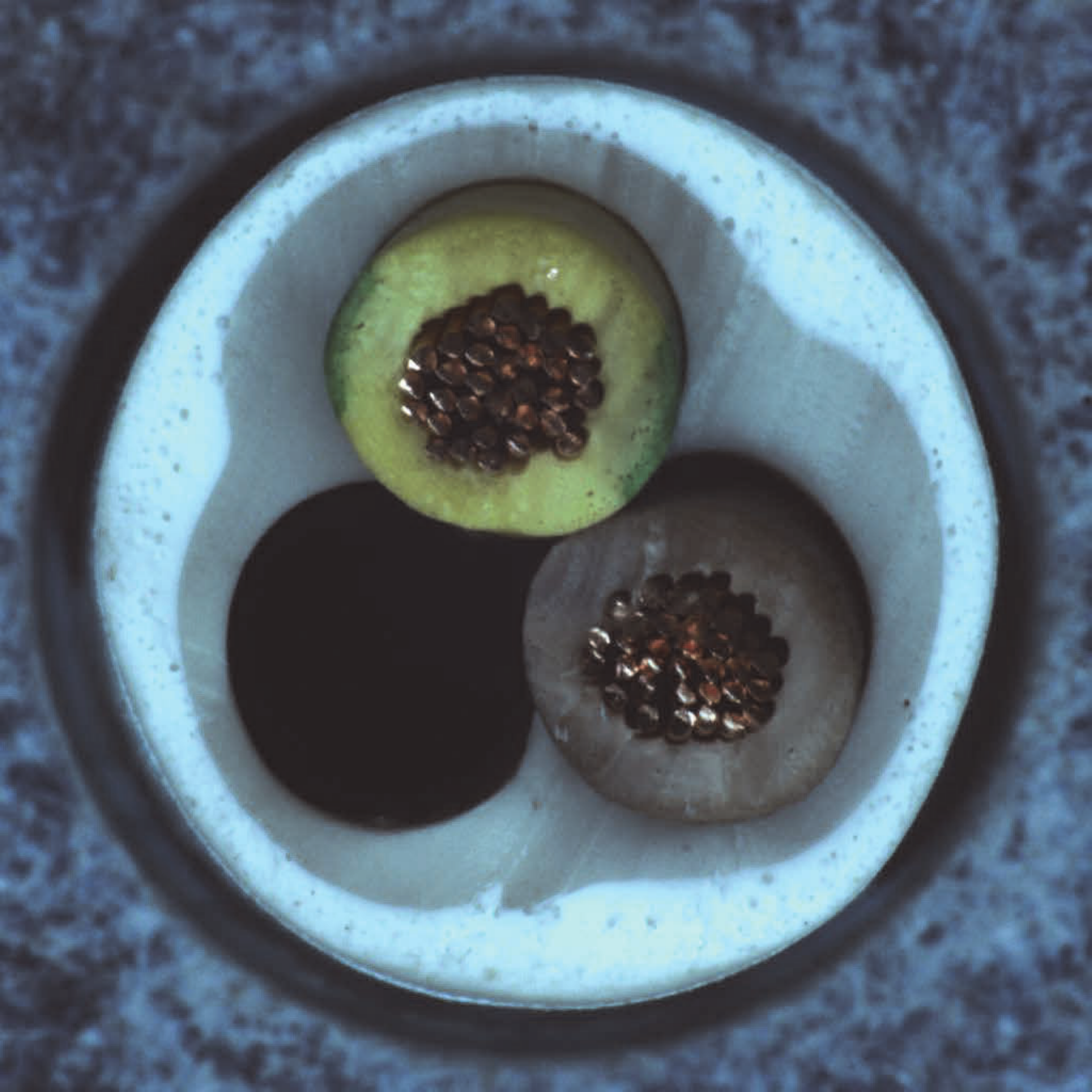}\\[1.5pt]
            \includegraphics[width=0.79in]{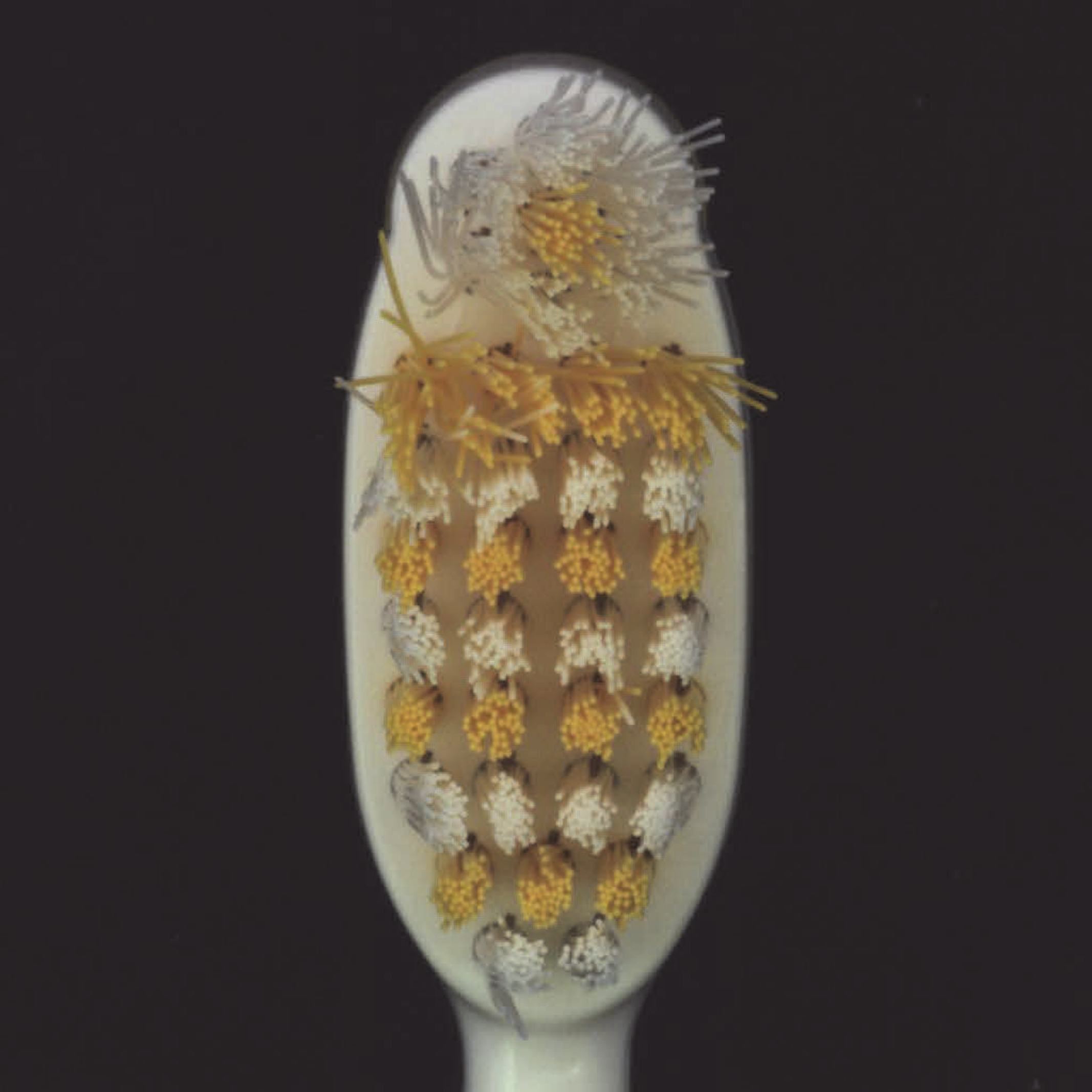}
        \end{tabular}%
	}
    \subfloat[ConvAE]{
        \begin{tabular}{@{}p{0.93in}@{}}\centering
			\includegraphics[width=0.79in]{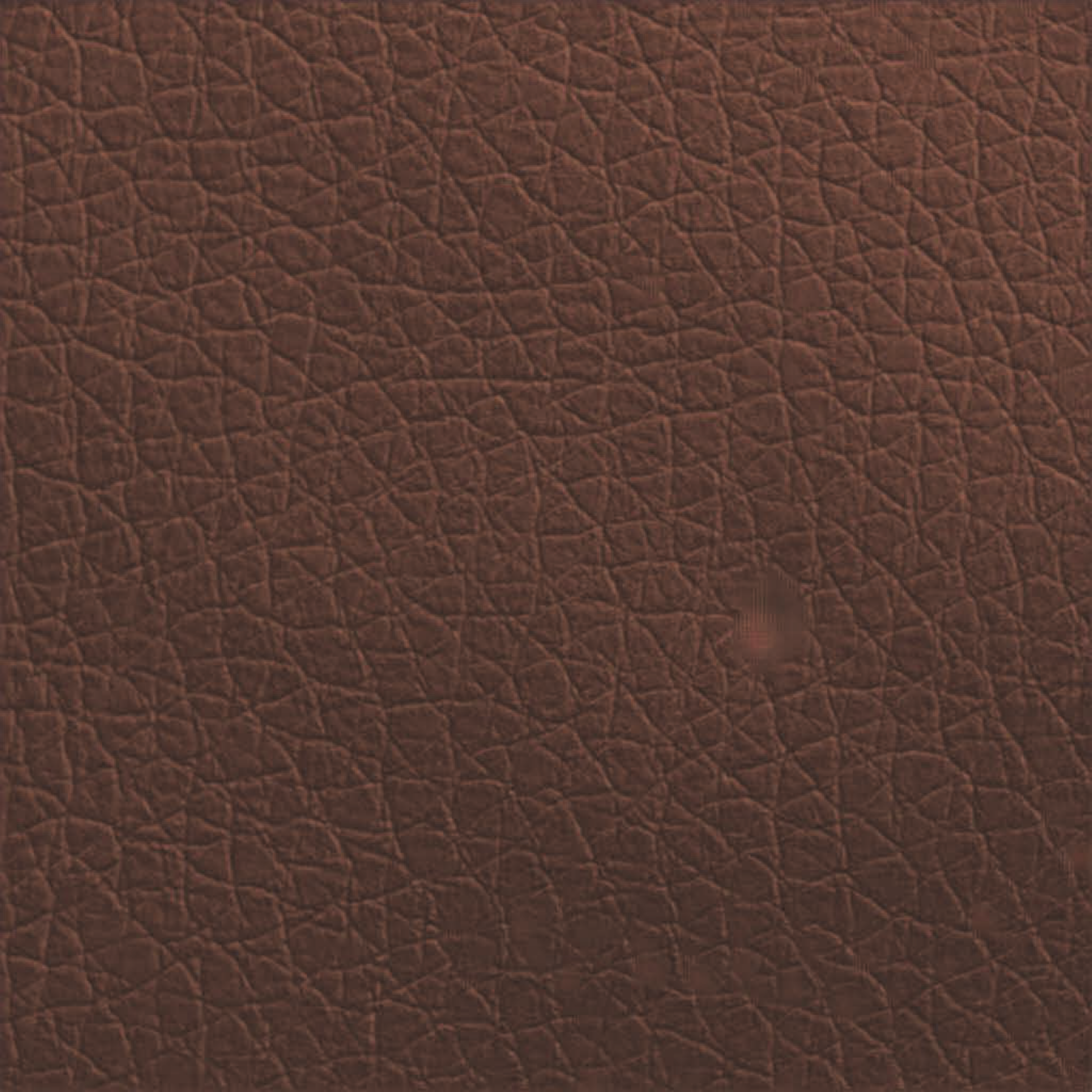}\\[1.5pt]
            \includegraphics[width=0.79in]{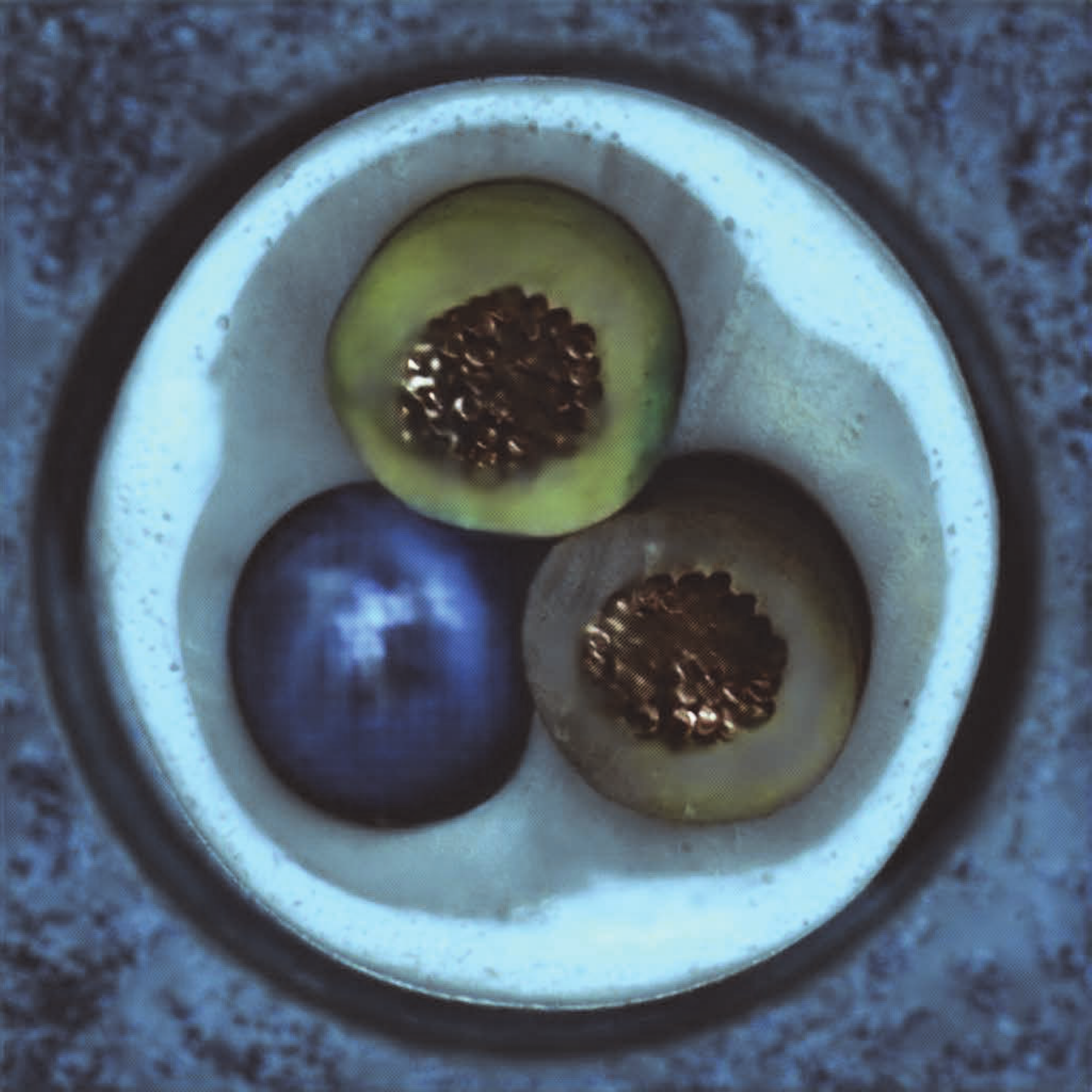}\\[1.5pt]
            \includegraphics[width=0.79in]{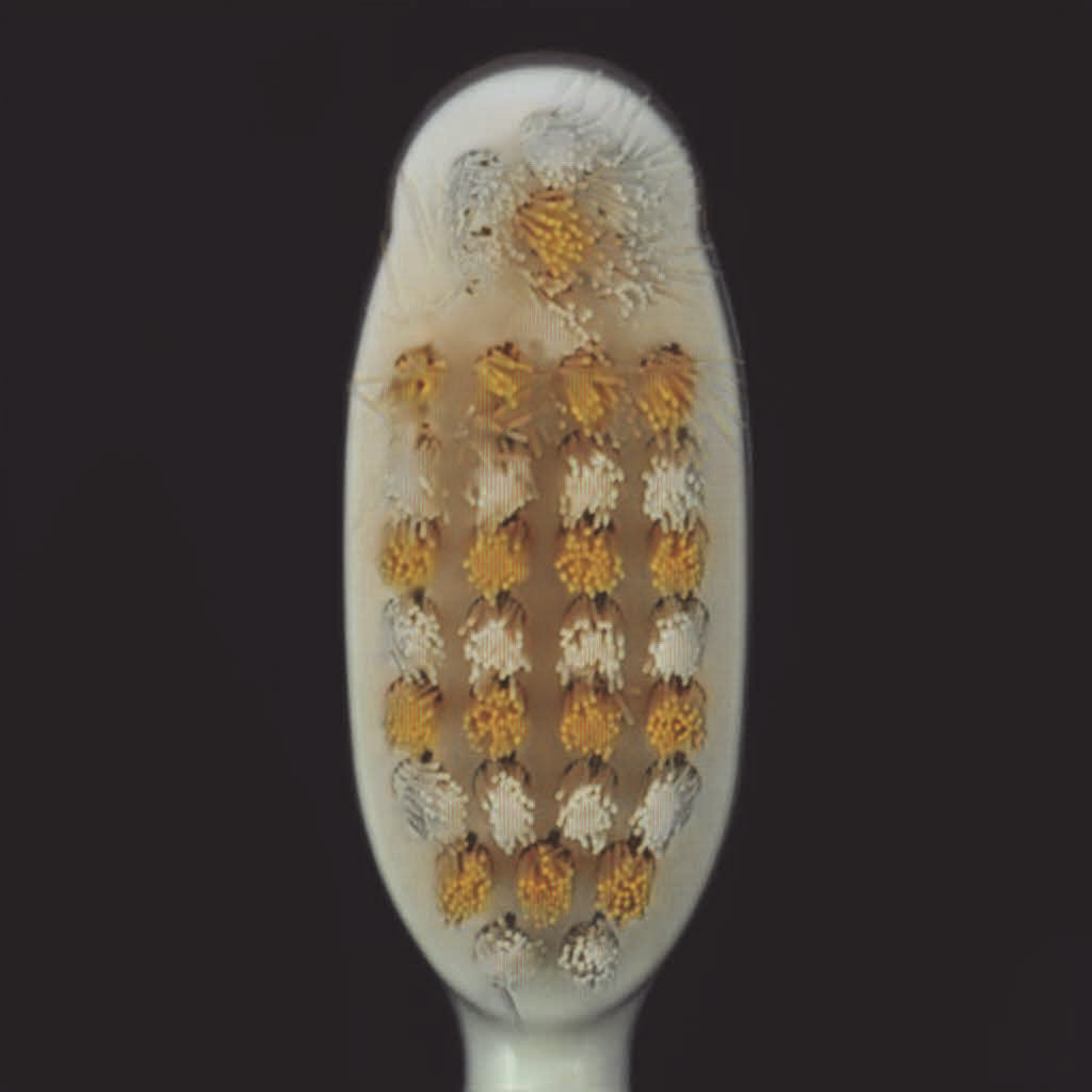}
        \end{tabular}%
	}
    \subfloat[RcAE w/o shared weight]{
        \begin{tabular}{@{}p{0.93in}@{}}\centering
			\includegraphics[width=0.79in]{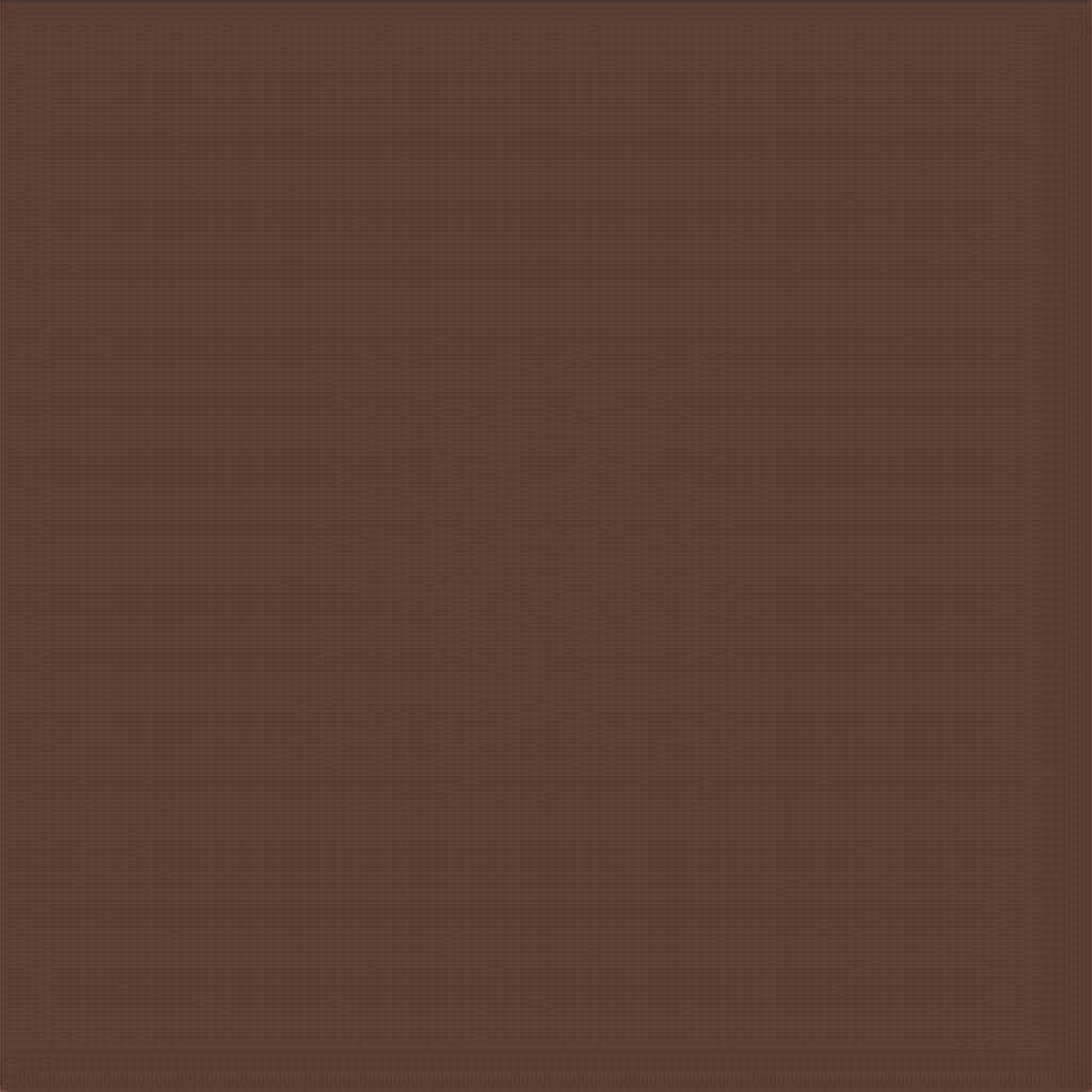}\\[1.5pt]
            \includegraphics[width=0.79in]{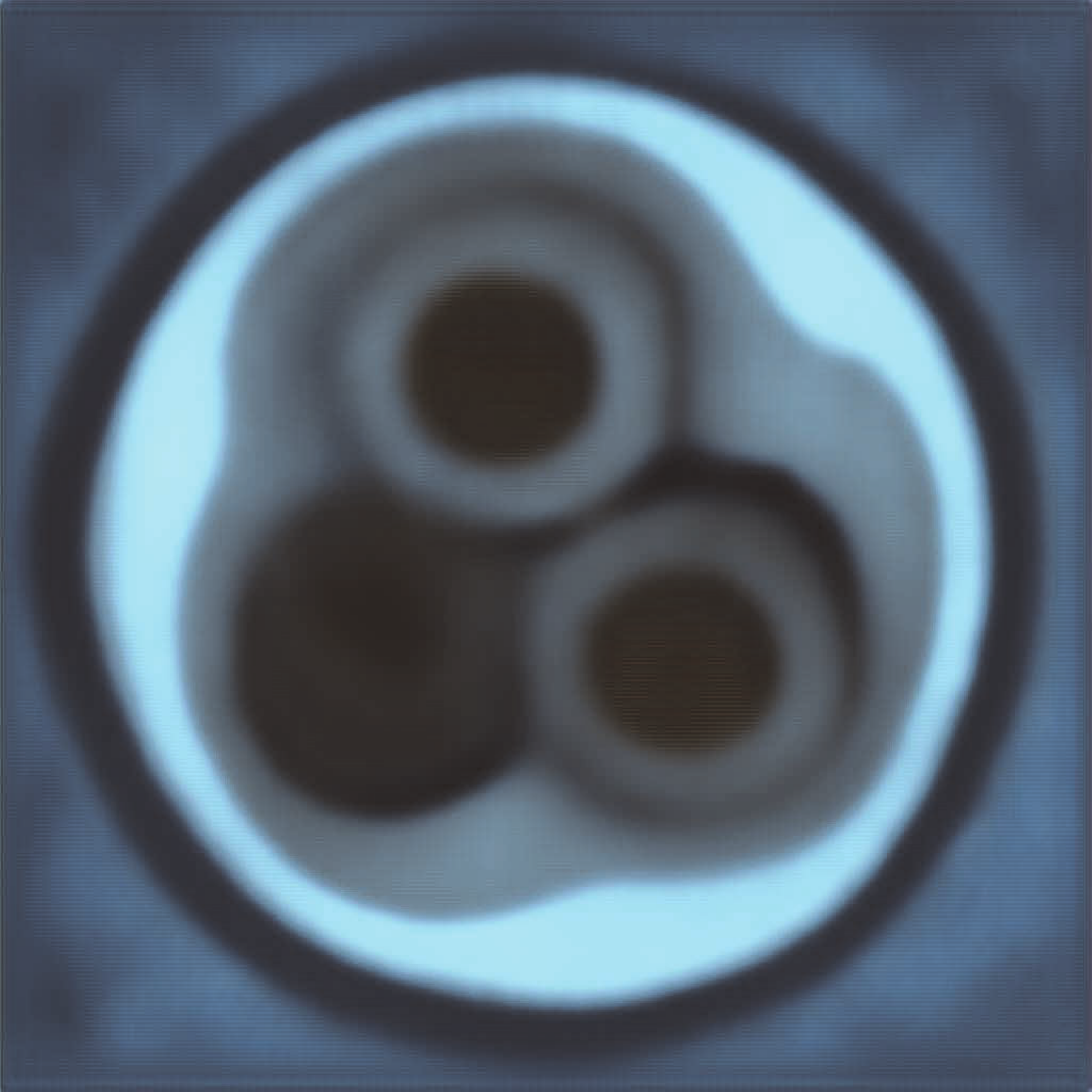}\\[1.5pt]
            \includegraphics[width=0.79in]{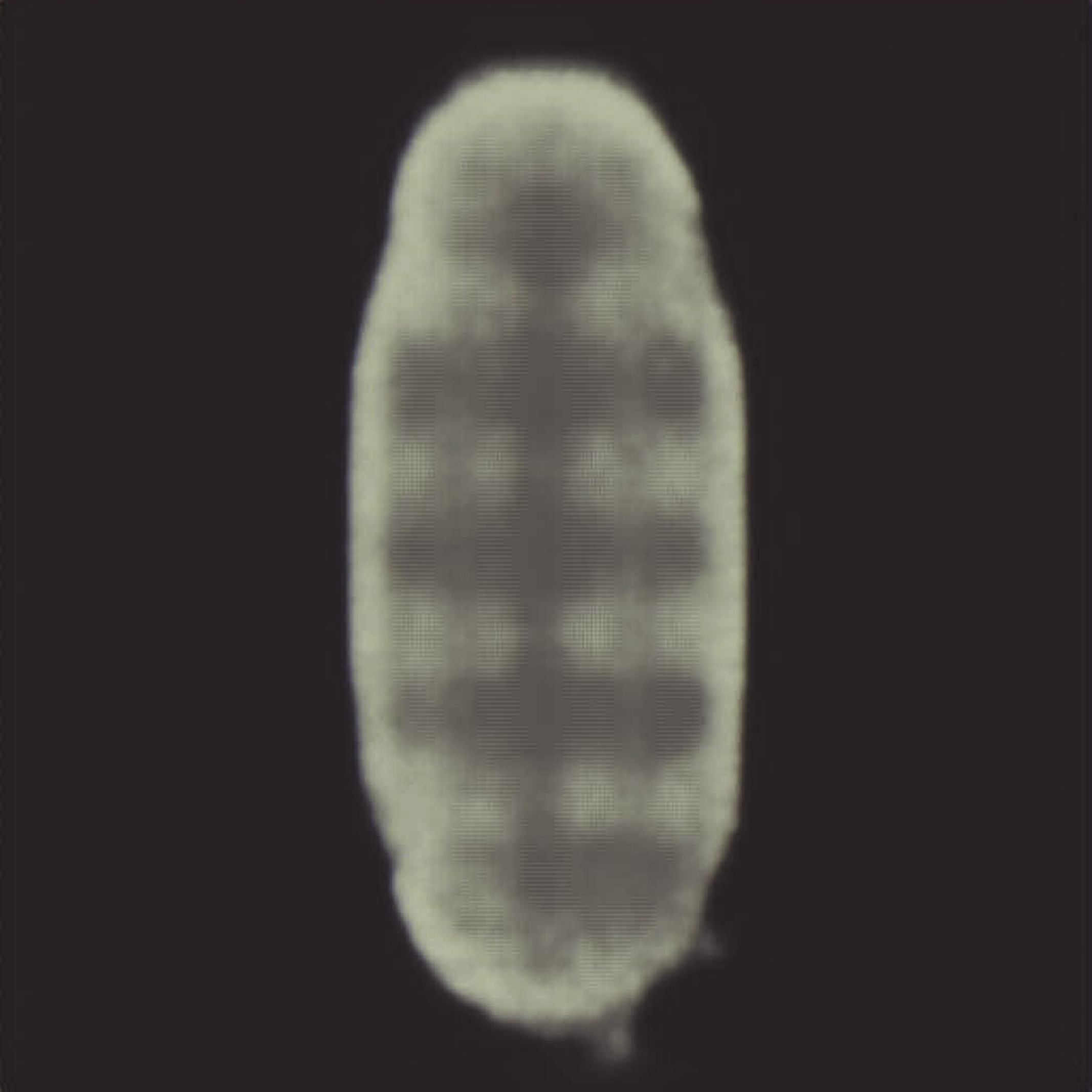}
        \end{tabular}%
	} 
    \subfloat[RcAE ($N$=3) w/o DPN]{
        \begin{tabular}{@{}p{0.93in}@{}}\centering
			\includegraphics[width=0.79in]{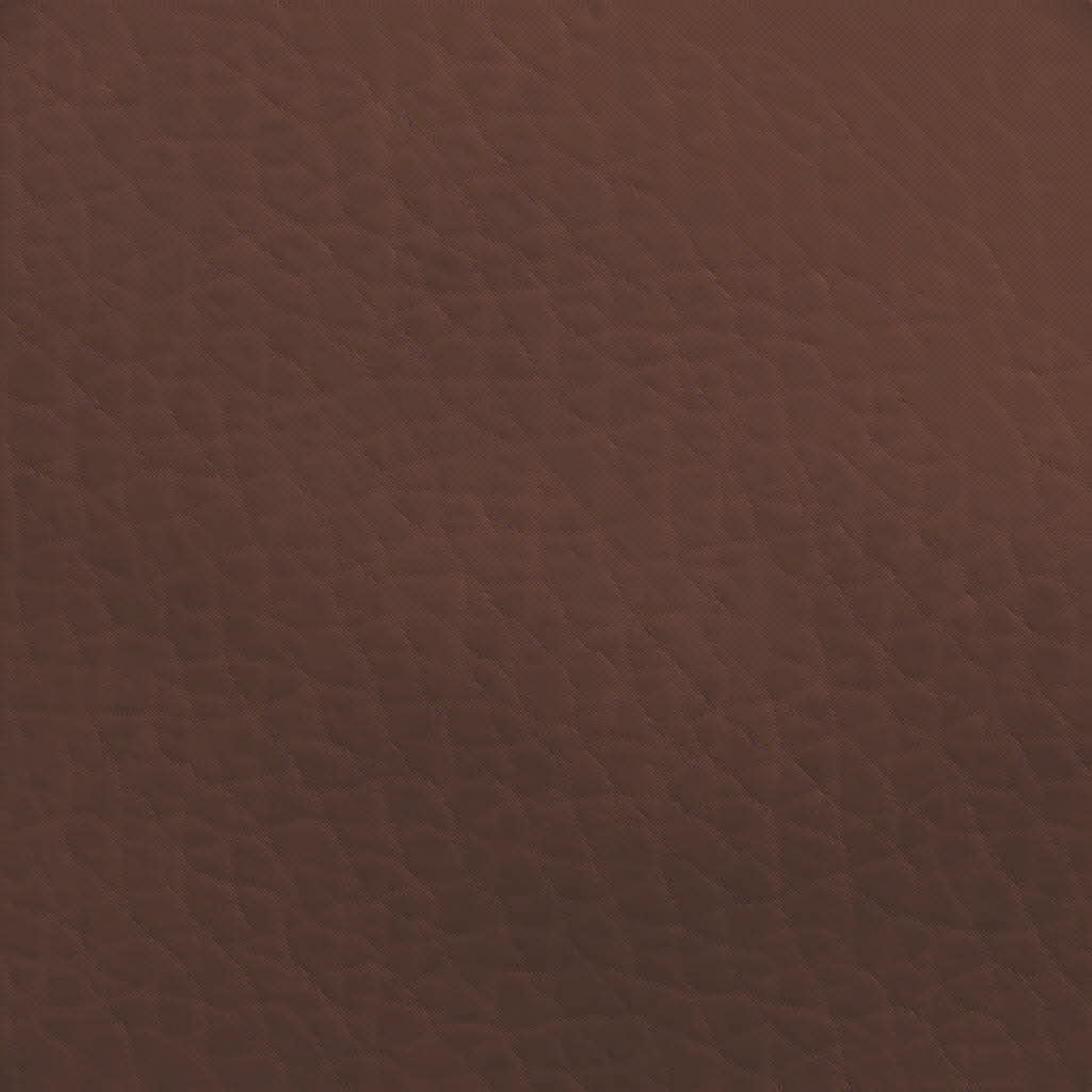}\\[1.5pt]
            \includegraphics[width=0.79in]{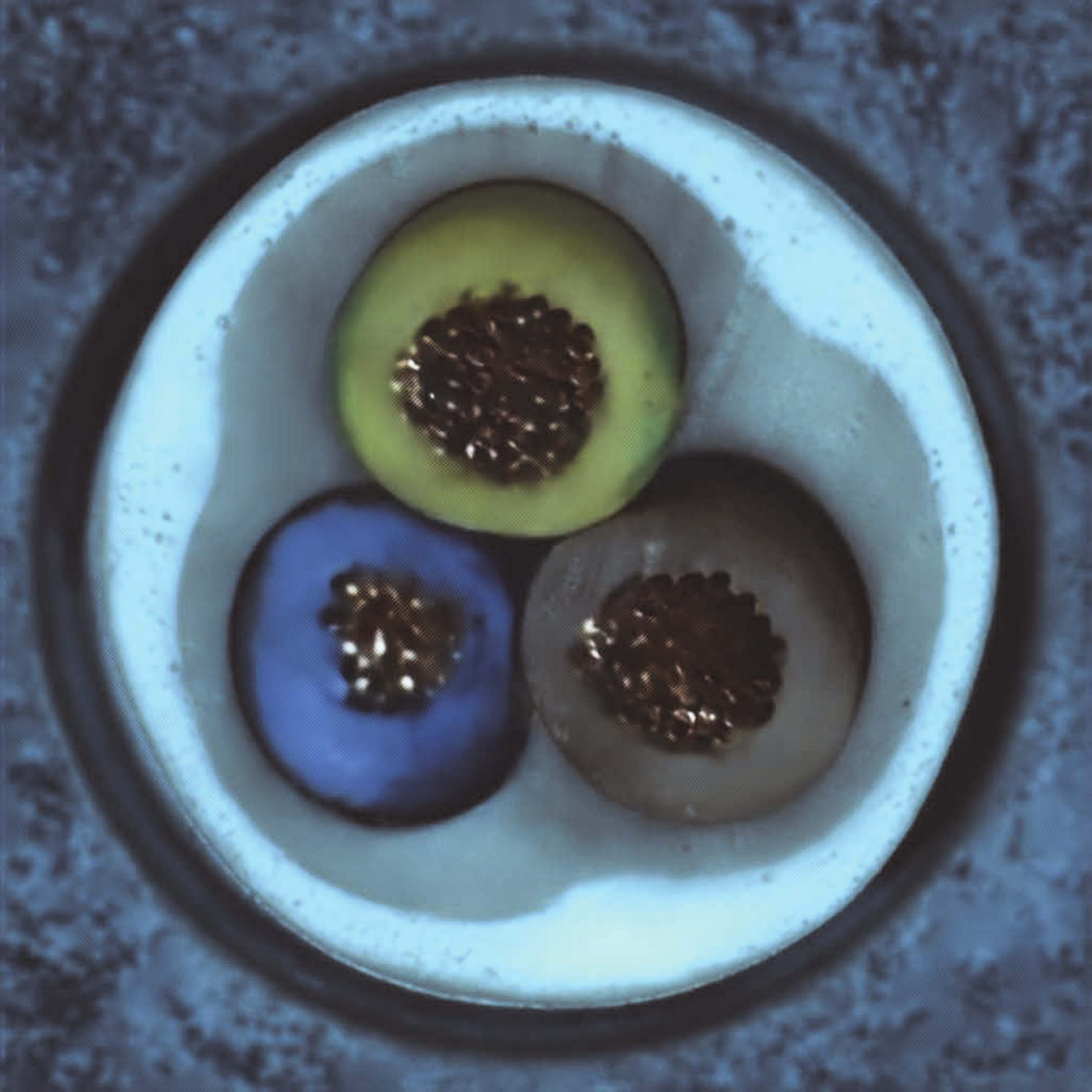}\\[1.5pt]
            \includegraphics[width=0.79in]{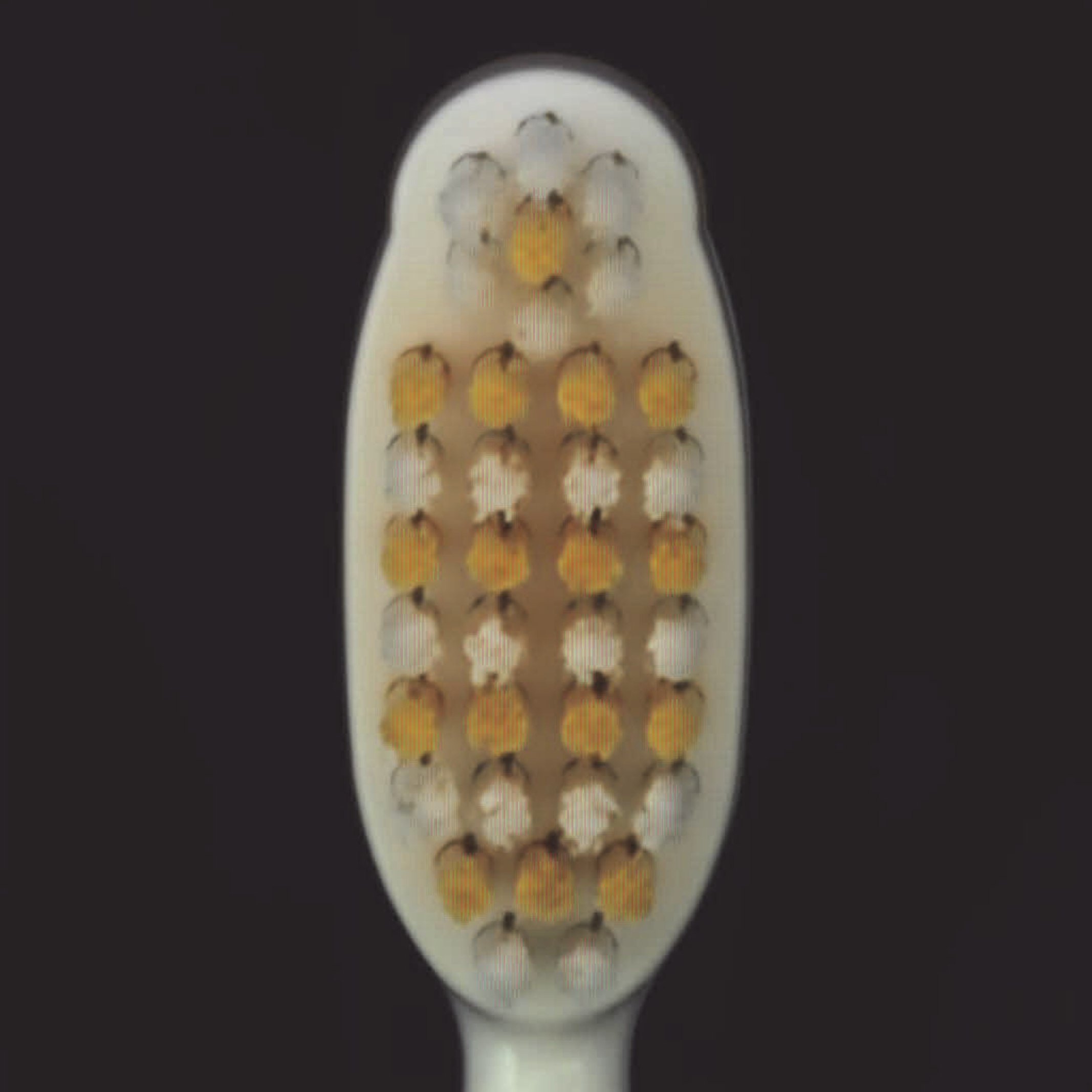}
        \end{tabular}%
	}    
	\subfloat[RcAE ($N$=5) w/o DPN]{
        \begin{tabular}{@{}p{0.93in}@{}}\centering
			\includegraphics[width=0.79in]{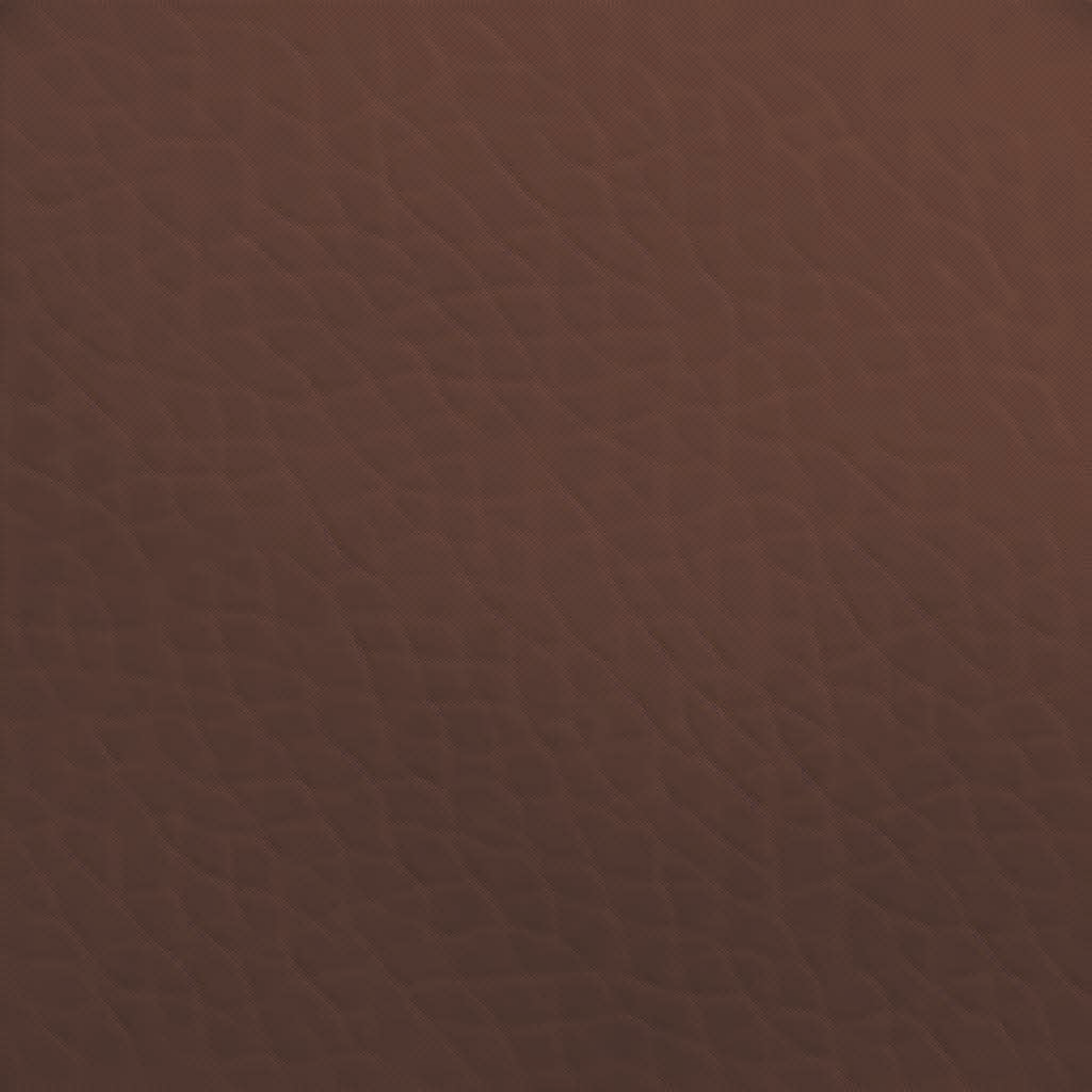}\\[1.5pt]
            \includegraphics[width=0.79in]{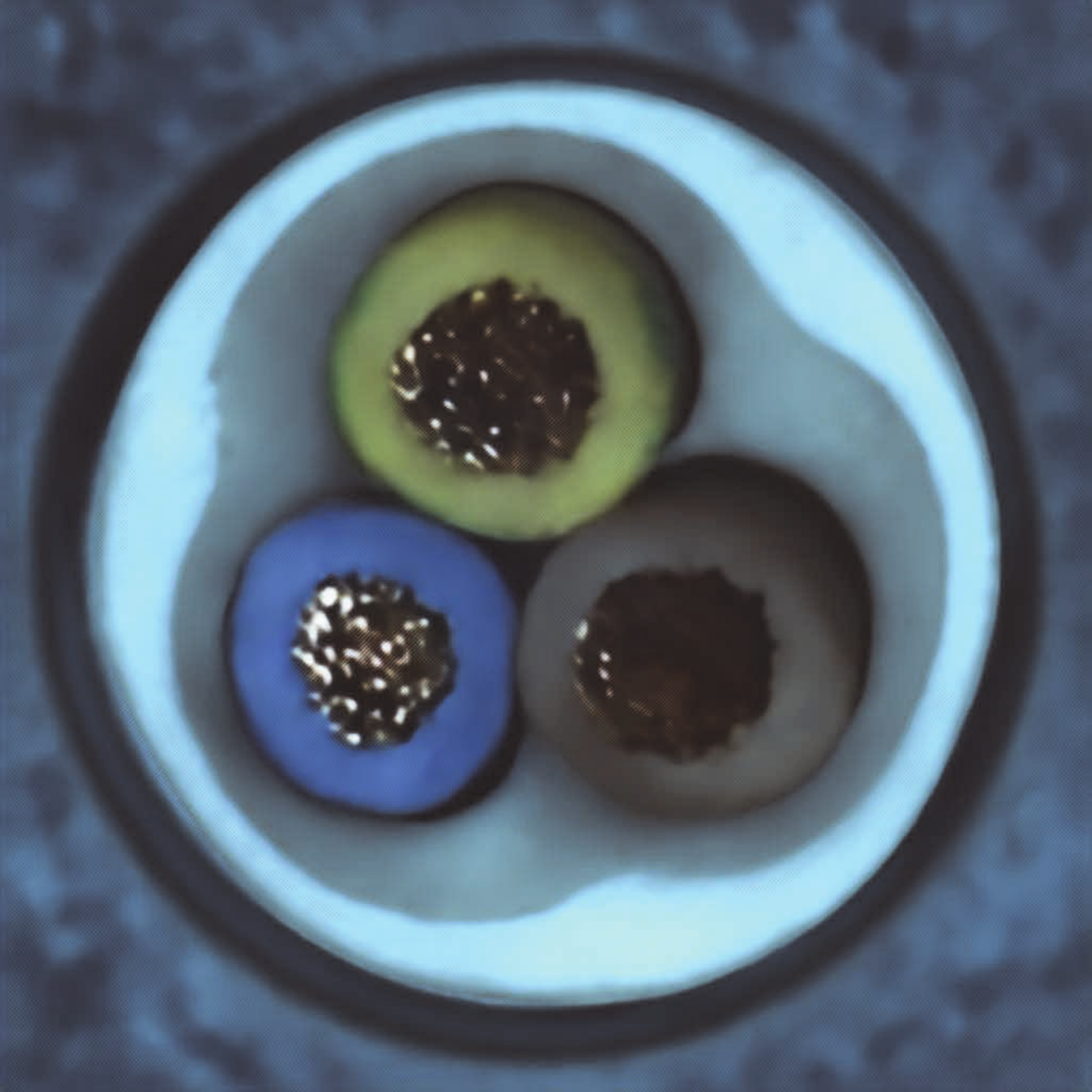}\\[1.5pt]
            \includegraphics[width=0.79in]{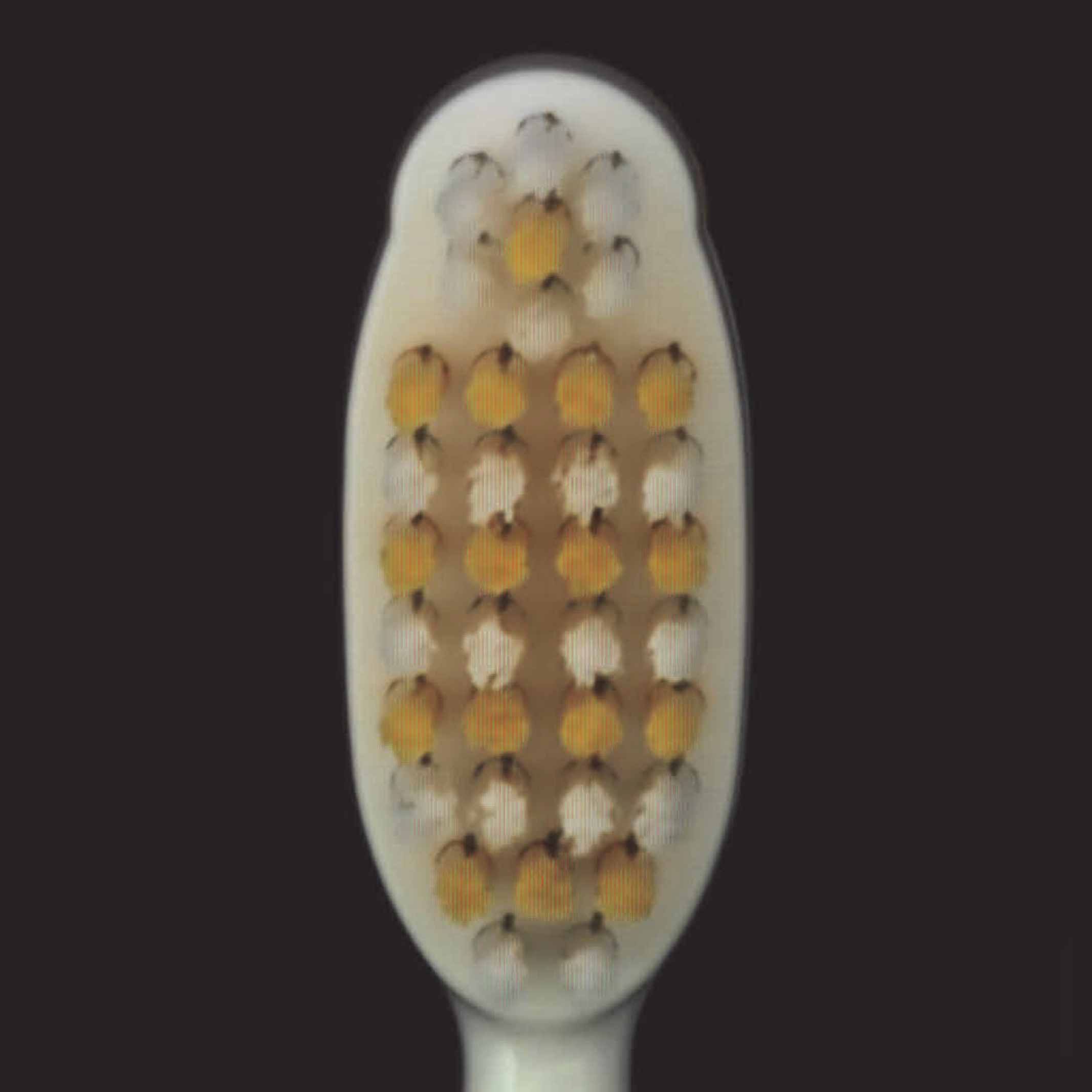}
        \end{tabular}%
	}
    \subfloat[RcAE ($N$=5)  w/ DPN]{
        \begin{tabular}{@{}p{0.93in}@{}}\centering
			\includegraphics[width=0.79in]{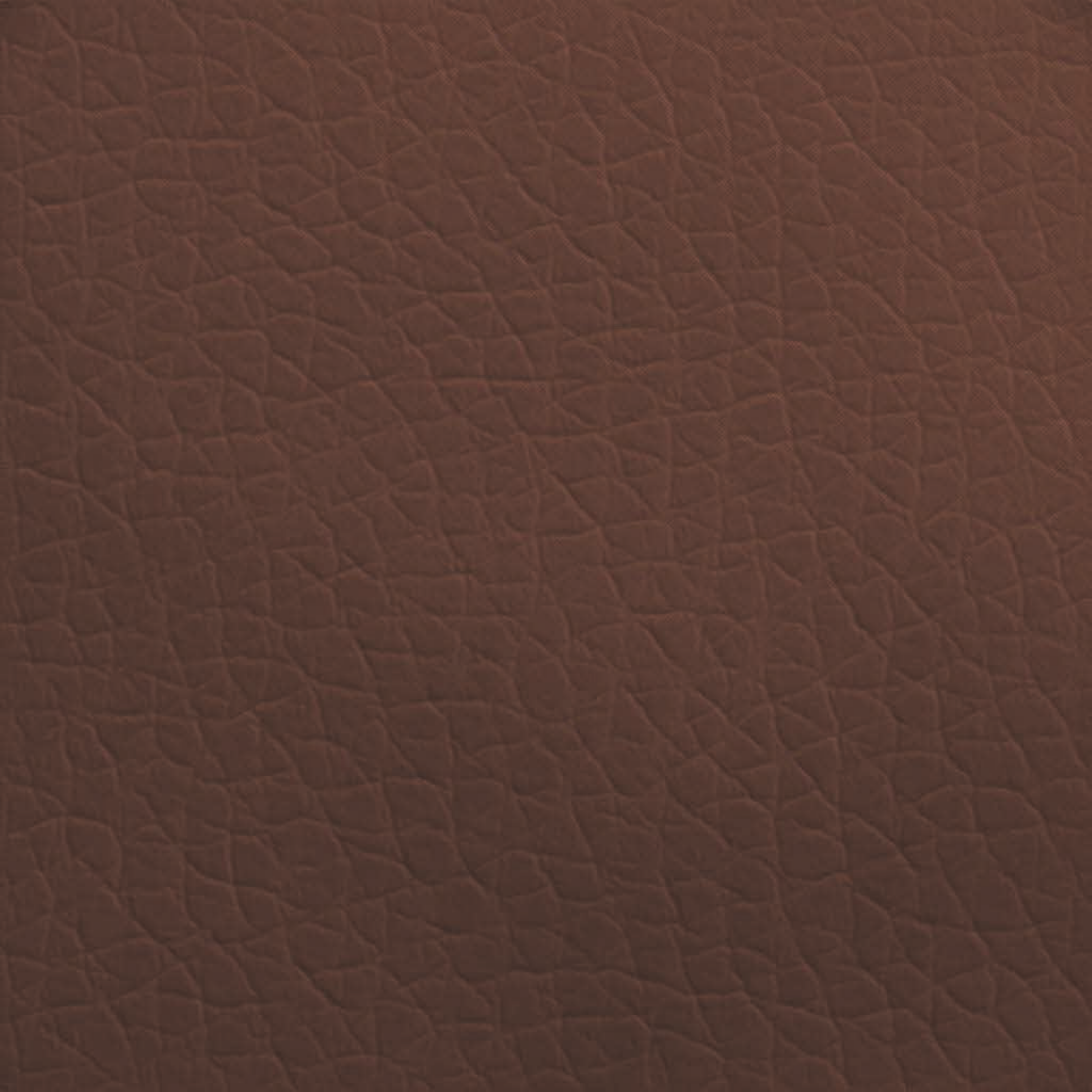}\\[1.5pt]
            \includegraphics[width=0.79in]{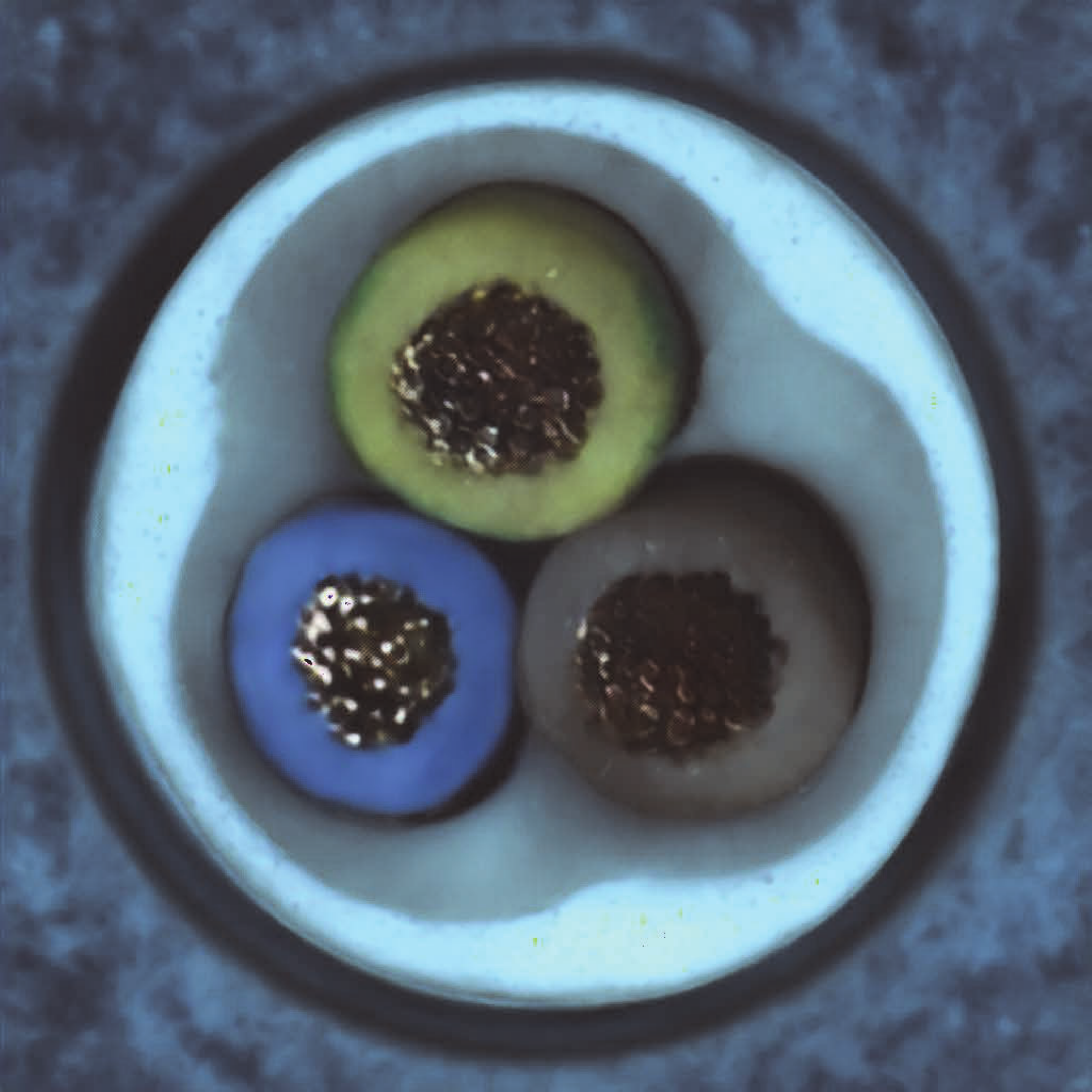}\\[1.5pt]
            \includegraphics[width=0.79in]{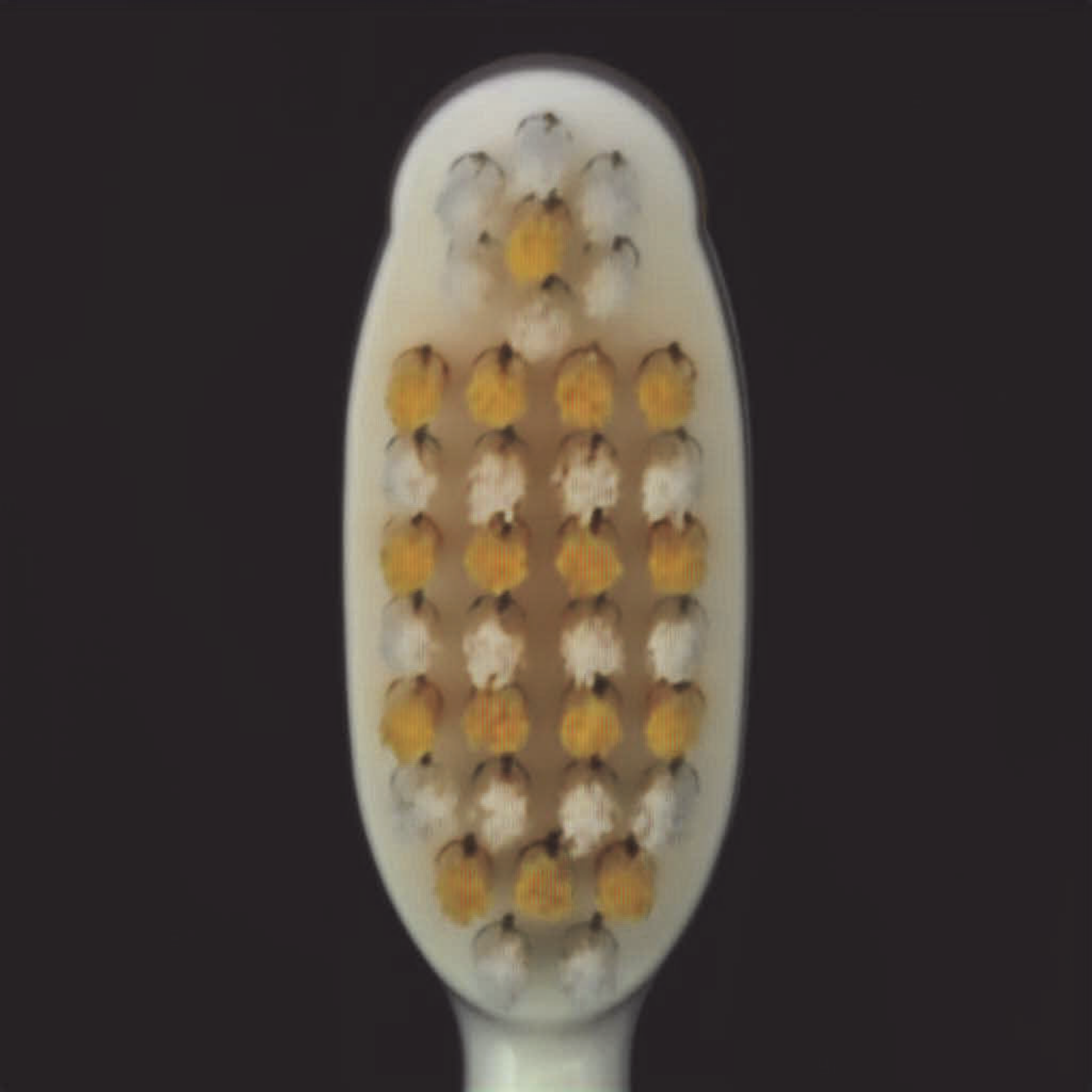}
        \end{tabular}%
	}
	\caption{Qualitative ablation on reconstruction. 
(a) Input. 
(b) ConvAE: blurry outputs with residual anomalies. 
(c) RcAE w/o weight sharing: partial anomaly memorization and detail loss. 
(d) RcAE ($N{=}3$): significantly improved reconstruction but anomalies remain. 
(e) RcAE ($N{=}5$): stronger suppression with slight texture erosion. 
(f) RcAE ($N{=}5$) + DPN: fine textures restored while anomalies remain suppressed.}
	\label{F_Ablation_ConvAE}
\end{figure*}

We comprehensively evaluate the proposed method on both image-level and pixel-level anomaly detection tasks, and analyze computational efficiency. Comparisons are made on the MVTec AD~\cite{MVTEC} and VisA~\cite{VisA} datasets with representative approaches across three categories:
(1) Non-diffusion methods: DRAEM~\cite{DREAM}, PatchCore~\cite{PatchCore}, RD4AD~\cite{RD4AD}, EfficientAD~\cite{Efficientad};
(2) Flow-based methods: MSFlow~\cite{msflow};
(3) Diffusion-based methods: D3AD~\cite{d3ad}, DiAD~\cite{Diad}, DiffAD~\cite{DiffAD}, and GLAD~\cite{GLAD}.



\vspace{0.05cm}
\noindent\textbf{Evaluation Metrics:}
Following common practice, we report image-level AUROC (I-AUROC) and pixel-level AUROC (P-AUROC) as primary evaluation metrics.

\vspace{0.05cm}
\noindent\textbf{Training Setup:} 
We train all components from scratch using Adam optimizer ($\eta$=$10^{-4}$, $\beta_1$=$0.9$, $\beta_2$=$0.999$, $\epsilon$=$10^{-8}$).
Stage 1/2/3 are trained for 1500/400/300 epochs, respectively, with recursion depth $N$=$5$ and input size 1024$\times$1024.
Experiments are conducted on NVIDIA RTX4090 with Python 3.10, and all baselines follow their official settings.


\subsection{Anomaly Detection and Localization}

Table~\ref{T_comparsion_exp} summarizes the anomaly detection and localization performance on the MVTec AD and VisA datasets. Our method consistently delivers strong results in both image and pixel level metrics while maintaining high efficiency.

On MVTec AD, our method achieves an average of 98.9\% I-AUROC and 98.7\% P-AUROC, outperforming non-diffusion baselines such as RD4AD (98.5/97.8\%) and DRAEM (98.0/97.3\%), and also surpassing diffusion models like D3AD (97.2/97.4\%), DiAD (97.2/96.8\%), and DiffAD (98.7/98.3\%). Notably, we achieve higher pixel-level accuracy than GLAD (98.7\% vs. 98.6\%), despite GLAD utilizing a large vision model (DINO~\cite{caron2021emerging}) with latent diffusion. This highlights the effectiveness of our pipeline, which achieves high performance without relying on pretrained models or heavy designs.

On the more challenging VisA dataset, which features complex object structures and diverse anomaly types, many methods exhibit performance degradation. Our method achieves 99.2\% I-AUROC and 98.6\% P-AUROC, tying GLAD for the second-best localization performance and ranking second in detection. It ranks among the top-2 in 10 out of 12 categories, highlighting strong robustness under realistic and challenging conditions.


Overall, our method achieves the \textbf{highest avg. P-AUROC (98.7\%)} and the \textbf{second-highest I-AUROC (99.0\%)} across both datasets. These results validate the effectiveness of our recursive reconstruction pipeline, highlighting our approach is highly suitable for practical industrial applications.

\subsection{Ablation Study}

\begin{table}[!t]
    \centering
    \begin{adjustbox}{max width=0.8\linewidth}
\begin{tabular}{@{}ccc|c@{}}
\toprule
~~~~~~RcAE~~~~~~ & ~~~~~~DPN~~~~~~ & ~~~~~~CRD~~~~ & ~~~~Accuracy~~~~ \\ \midrule
\ding{55}    & \ding{55}    & \ding{55}     & 82.4/90.8 \\
\ding{51}    & \ding{55}    & \ding{55}     & 94.1/95.8 \\
\ding{51}    & \ding{51}    & \ding{55}     & 95.7/96.6 \\
\ding{51}    & \ding{51}    & \ding{51}     & \textbf{98.9/98.7}  \\ \bottomrule
\end{tabular}
\end{adjustbox}
\caption{Ablation on core components of our method. Each entry shows I-AUROC / P-AUROC (\%) on MVTec AD.}
\label{T_Ablation_Core}
\end{table}

\noindent\textbf{Effectiveness of Core Components.}
Table~\ref{T_Ablation_Core} presents a step-wise ablation. Starting from a plain ConvAE baseline (82.4/90.8\% I-/P-AUROC), introducing RcAE already lifts performance to 94.1/95.8\% by progressively suppressing anomalies and refining normal structures.
Adding the DPN improves results to 95.7/96.6\% by selectively restoring high-frequency details that may be smoothed during recursive reconstruction.
Finally, integrating CRD yields the full model with 98.9/98.7\%, leveraging cross-recursion residual dynamics for robust anomaly localization.
Fig.~\ref{F_Ablation_ConvAE} provides qualitative evidence: ConvAE outputs are blurry with residual defects, RcAE progressively normalizes anomalies, and DPN restores textures without reintroducing them.

\vspace{0.1cm}
\noindent\textbf{Impact of Recursion Depth of RcAE.}
We analyze the effect of recursion depth $N$ in Table~\ref{T_rcAE_iteration}.
At $N$=1, RcAE equals to two ConvAEs with limited capacity (86.2/87.4\%). Increasing $N$ strengthens anomaly suppression and semantic refinement, with the largest gain between $N$=1 and $N$=3, and a peak at $N$=5.
Beyond this, performance plateaus or slightly declines, suggesting diminishing returns and minor over-smoothing.
This validates the benefit of recursive architecture in reconstruction quality and exposing anomalies. Fig.~\ref{F_Ablation_ConvAE} illustrates these trends qualitatively, and more reconstruction results can be seen in Supplementary Materials.

\begin{table}[!h]
\centering
\begin{adjustbox}{max width=1\linewidth}
\begin{tabular}{@{}lcccccc@{}}
\toprule
\multirow{2}{*}{Dataset} & \multicolumn{6}{c}{Number of Iterations ($N$)}                                 \\ \cmidrule(l){2-7} 
                         & $N$=1     & $N$=2     & $N$=3     & $N$=4     & $N$=5              & $N$=6     \\ \midrule
MVTec                 & 86.2/87.4 & 90.8/89.9 & 96.3/96.8 & 98.0/98.3 & \textbf{98.9/98.7} & 98.7/98.4 \\
VisA                     & 89.3/88.7 & 93.2/92.1 & 97.3/96.4 & 98.9/98.1 & \textbf{99.2/98.6} & \textbf{99.2}/98.1 \\ \bottomrule
\end{tabular}
\end{adjustbox}
\caption{Effect of RcAE recursion depth $N$ on anomaly detection (I-AUROC / P-AUROC (\%)).}
\label{T_rcAE_iteration}
\end{table}

\vspace{0.1cm}
\noindent\textbf{Impact of RcAE Architecture.}
Table~\ref{T_rcAE_arch} shows how skip connections and parameter sharing affect RcAE performance. Replacing the basic ConvAE (65.2/72.4\%) with a recursive design significantly improves accuracy, and skip connections further boosts it to 98.9/98.7\%. Conversely, removing weight sharing drops performance to 71.3/74.2\%, highlighting the importance of both recursion and constrained parameterization.
Notably, for anomaly detection, skip connections often lead to shortcut learning in ConvAE. In RcAE, the repeated compression–reconstruction suppress shortcuts, while skip connections enhance shallow feature propagation, improving reconstruction quality.



\begin{table}[!h]
\centering
\begin{adjustbox}{max width=0.95\linewidth}
\begin{tabular}{@{}lccc c@{}}
\toprule
\multirow{2}{*}{Model} & \multirow{2}{*}{Skip Conn.} & \multirow{2}{*}{Shared Weights} & \multicolumn{2}{c}{Accuracy} \\ \cmidrule(l){4-5}
                       &                             &                                & MVTec AD        & VisA        \\ \midrule
ConvAE                 & \ding{51}                   & \ding{55}                       & 65.2 / 72.4     & 68.5 / 70.1 \\
ConvAE                 & \ding{55}                   & \ding{55}                       & 82.4 / 90.8     & 79.7 / 78.5 \\
RcAE                   & \ding{55}                   & \ding{51}                       & 92.2 / 95.2     & 94.3 / 95.9 \\
RcAE                   & \ding{51}                   & \ding{55}                       & 71.3 / 74.2     & 73.5 / 75.0 \\
\textbf{RcAE}   & \ding{51}                   & \ding{51}                       & \textbf{98.9 / 98.7} & \textbf{99.2 / 98.6} \\ 
\bottomrule
\end{tabular}
\end{adjustbox}
\caption{Effect of skip connections and parameter sharing in RcAE (I-AUROC / P-AUROC (\%)).}
\label{T_rcAE_arch}
\end{table}



\noindent\textbf{Data Efficiency.}
Table~\ref{T_data_size} shows that RcAE consistently outperforms ConvAE across all training data ratios.
With only 10\% of the training data, RcAE already surpasses the full-data ConvAE, demonstrating superior data efficiency, which is important for industrial scenarios with scarce samples.

\begin{table}[!h]
\centering
\begin{adjustbox}{width=0.48\textwidth}
\begin{tabular}{lccccc}
\toprule
\multirow{2}{*}{Method} & \multicolumn{5}{c}{Training Data Percentage}              \\ \cmidrule(l){2-6} 
                        & 10\%      & 25\%      & 50\%      & 75\%      & 100\%     \\ \midrule
ConvAE                  & 62.3/61.5 & 68.9/77.2 & 71.4/87.8 & 79.1/88.9 & 82.4/90.8 \\
\textbf{RcAE}    & \textbf{84.1/93.4} & \textbf{91.2/93.7} & \textbf{95.5/95.1} & \textbf{97.2/96.8} & \textbf{98.9/98.7} \\
\bottomrule
\end{tabular}
\end{adjustbox}
\caption{Impact of training data size on anomaly detection performance on MVTec AD (I-AUROC / P-AUROC (\%)).}
\label{T_data_size}
\end{table}

\noindent\textbf{Effectiveness of Detail Preservation Network.}
Detail Preservation Network mitigates detail loss that can lead to false positives. Table~\ref{T_Ablation_Core} shows that adding DPN improves I-/P-AUROC from 94.1/95.8\% to 95.7/96.6\%. It also improves reconstruction quality with average SSIM and PSNR gains of 0.059 and 0.61 dB (up to 0.32 and 2.61 dB). 

As shown in Fig.~\ref{F_Ablation_ConvAE}(e-f), DPN selectively restores details while maintaining anomaly suppression, especially on highly textured samples (e.g., leather), confirming that DPN preserves details without reintroducing anomalies.

\begin{table}[!h]
\centering
\begin{adjustbox}{max width=0.77\linewidth}
\begin{tabular}{@{}c|c|c@{}}
\toprule
$N_{\text{R}}$ & Recon steps fed to CRD & I-AUROC / P-AUROC \\ \midrule
1 & \{5\}         & 95.7 / 96.6 \\
3 & \{1, 3, 5\}   & 98.0 / 97.1 \\
\textbf{5} & \textbf{\{1–5\}}       & \textbf{98.9 / 98.7} \\ 
\bottomrule
\end{tabular}
\end{adjustbox}
\caption{Impact of the number of RcAE reconstructions used by CRD (I-AUROC / P-AUROC (\%) on MVTec AD).}
\label{T_differentN_MSD}
\end{table}

\vspace{0.1cm}
\noindent\textbf{Effectiveness of Cross Recursion Detection.}
We fix $N{=}5$ in RcAE and vary the number of reconstructions $N_{\text{R}}$ fed into CRD (Table~\ref{T_differentN_MSD}).
Using only the final reconstruction ($N_{\text{R}}{=}1$, step 5) provides a strong baseline (95.7/96.6\%).
Incorporating intermediate reconstructions ($N_{\text{R}}{=}3$, steps 1/3/5) improves results to 98.0/97.1\%, and using all steps ($N_{\text{R}}{=}5$, steps 1–5) achieves 98.9/98.7\%.
This demonstrates that leveraging reconstruction dynamics across recursion steps yields more reliable anomaly localization than a single residual map.


\subsection{Computational Complexity}
\label{sec:cost}

\begin{figure}[!h]
    \centering
    \includegraphics[width=1\linewidth]{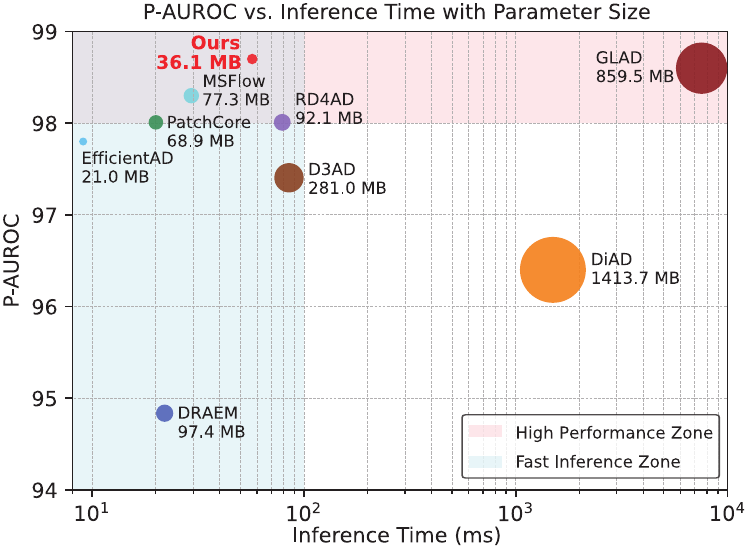}
    \caption{Trade-off between computation cost and accuracy. Circle size indicates model parameter count.}
    \label{F_Cost}
\end{figure}

As shown in Fig.~\ref{F_Cost}, our method achieves a favorable balance between accuracy and efficiency. The recursive architecture slightly increases inference time compared to single-pass ConvAEs , but it maintains a compact parameter count with high performance and remains much faster than diffusion. Despite this lightweight design without pretraining or external priors, the accuracy is on par with GLAD, which requires both latent diffusion and DINO. This combination of high performance, small size, and good latency makes our method practical for real-world applications.



\section{Conclusion}

We proposed an efficient recursive autoencoder framework for industrial anomaly detection. 
By sharing parameters across iterations, the model progressively refines reconstructions without increasing model size and leverages cross-recursion dynamics for robust anomaly localization. 
Trained entirely from scratch without external priors, our method achieves state-of-the-art performance with practical computational efficiency.  
Despite these advantages, the current design may be less effective for high-level logical anomalies that require semantic reasoning. 
Future work will explore integrating lightweight prior knowledge or hybrid architectures to address such cases, and extending the recursive paradigm to broader industrial vision tasks facing similar efficiency and data limitations.

\section*{Acknowledgments}
This work was supported by the National Natural Science Foundation of China (No.~62306223), the China Postdoctoral Science Foundation (No.~2024M752533), the Young Talent Fund of Xi'an Association for Science and Technology (No.~959202413053), the Fundamental Research Funds for the Central Universities (No.~XJSJ24020), the Postdoctoral Science Foundation of Shaanxi Province, the Xiaomi Young Talents Program, and the Digital Finance CRC (supported by the Cooperative Research Centres program, an Australian Government initiative).

\bibliography{aaai2026}

\clearpage
\appendix


\section*{Supplementary material}

# Supplementary Material for
# RcAE: Recursive Reconstruction Framework for Unsupervised Industrial Anomaly Detection

## Theoretical Foundations of RcAE

To theoretically support the design choices and superior performance demonstrated by our Recursive Convolutional Autoencoder (RcAE), we present an in-depth analysis highlighting how its recursive design and parameter-sharing strategy contribute to distinct advantages in function space constraints, generalization performance, anomaly suppression, and multi-scale information capacity.

As introduced in Section 3 of the main paper, our recursive architecture repeatedly leverages shared encoder and decoder modules. Below, we rigorously elucidate how this architecture inherently provides beneficial inductive biases and theoretical guarantees, which align closely with the experimental observations presented in Sections 4.

### Scale-Space Analysis

From a scale-space theoretical perspective, RcAE implements hierarchical processing of input images. Traditional deep autoencoder architectures typically implement multi-scale feature processing using distinct parameters for each scale. As empirically shown in our ablation studies (Table 4), such a design can result in overfitting to scale-specific features, causing undesirable anomaly memorization effects and limited generalization. In contrast, our recursive architecture implements a hierarchical analysis of the input image via shared encoder and decoder functions applied recursively.

Let $\mathcal{T}_\sigma : \mathcal{X} \to \mathcal{X}$ represent a transformation that processes features at scale $\sigma$. In conventional deep networks with encoder-decoder architecture, these transformations are implemented with distinct parameters:

$$\mathcal{T}_{\sigma_i}(\mathbf{x}) = D_i \circ D_{i+1} \circ ... \circ D_N \circ E_N \circ ... \circ E_{i+1} \circ E_i(\mathbf{x}),$$
$$i \in \{1, 2, ..., N\}, \quad (1)$$

where each encoder block $E_i$ and decoder block $D_i$ has its own parameters $\boldsymbol{\theta}_{E_i}$ and $\boldsymbol{\theta}_{D_i}$, respectively. In contrast, our recursive approach implements multi-scale analysis through repeated application of shared encoder and decoder functions:

$$\mathcal{T}_{\sigma_n}(\mathbf{x}) = D^{(n)}(E^{(n)}(\mathbf{x})), \quad n \in \{1, 2, ..., N\}, \quad (2)$$

where $E^{(n)}$ represents $n$ recursive applications of our shared encoder with shared parameters $\boldsymbol{\theta}_E$, and $D^{(n)}$ represents $n$ recursive applications of our shared decoder with shared parameters $\boldsymbol{\theta}_D$. This parameter sharing enforces scale-consistency constraints through gradient accumulation during backpropagation:

$$\frac{\partial \mathcal{T}_{\sigma_n}(\mathbf{x})}{\partial \boldsymbol{\theta}_E} = \sum_{i=1}^{n} \frac{\partial D^{(n)}(E^{(n)}(\mathbf{x}))}{\partial E^{(i)}(\mathbf{x})} \cdot \frac{\partial E^{(i)}(\mathbf{x})}{\partial \boldsymbol{\theta}_E}, \quad (3)$$

and

$$\frac{\partial \mathcal{T}_{\sigma_n}(\mathbf{x})}{\partial \boldsymbol{\theta}_D} = \sum_{j=1}^{n} \frac{\partial D^{(j)}(E^{(n)}(\mathbf{x}))}{\partial \boldsymbol{\theta}_D}. \quad (4)$$

During backpropagation, this gradient accumulation across scales forces the network to learn features that are consistent across multiple scales rather than scale-specific features that might enable anomaly memorization.

This theoretical property is directly reflected in our practical experiments (Tables 2 and 3), this multi-scale processing directly translates to effective detection of diverse anomalies: early iterations ($n = 1$) detect local defects (e.g., surface scratches), while later iterations ($n = 5$) capture global structural anomalies (e.g., missing components).

### Function Space Constraints

To further analyze the effects of parameter sharing, we formally establish in Theorem 1 that RcAE inherently operates in a more restricted function space compared to conventional deep autoencoders. Such a constraint provides a strong inductive bias towards learning scale-consistent transformations, which is especially beneficial for anomaly detection.

In Section 4, our experiments clearly demonstrate that conventional deep autoencoders without recursive constraints suffer from severe anomaly memorization, reflected by inferior anomaly detection metrics. Conversely, RcAE's constrained function space significantly mitigates this issue.

We rigorously support this observation with the following theorem:

**Theorem 1** (Function Space Constraint). *Let $\mathcal{F}_{deep} = \{f_{deep} : \mathcal{X} \to \mathcal{X} \mid f_{deep}(\mathbf{x}) = D_1 \circ D_2 \circ ... \circ D_N \circ E_N \circ ... \circ E_2 \circ E_1(\mathbf{x})\}$ be the set of functions representable by conventional deep autoencoders with $N$ distinct encoder-decoder pairs, and $\mathcal{F}_{rec} = \{f_{rec} : \mathcal{X} \to \mathcal{X} \mid f_{rec}(\mathbf{x}) = D^{(N)}(E^{(N)}(\mathbf{x}))\}$ be*

*the set of functions representable by our recursive architecture with shared encoder $E$ and decoder $D$ applied $N$ times. Then:*

$$\mathcal{F}_{rec} \subset \mathcal{F}_{deep}. \quad (5)$$

This theorem establishes that the recursive architecture operates in a strictly smaller function space than conventional architectures, limiting its capacity to memorize arbitrary patterns.

*Proof. (Theorem 1)* Let $f_{\text{rec}} \in \mathcal{F}_{\text{rec}}$ be any function representable by our recursive architecture. By definition, $f_{\text{rec}}(\mathbf{x}) = D^{(N)}(E^{(N)}(\mathbf{x}))$, where $E$ and $D$ are functions parameterized by shared parameters $\boldsymbol{\theta}_E$ and $\boldsymbol{\theta}_D$, respectively.

We can construct an equivalent function $f_{\text{deep}} \in \mathcal{F}_{\text{deep}}$ by setting:

$$\begin{aligned} E_1 = E_2 = \ldots = E_N = E \\ D_1 = D_2 = \ldots = D_N = D \end{aligned} \quad (6)$$

with identical parameters $\boldsymbol{\theta}_{E_i} = \boldsymbol{\theta}_E$ and $\boldsymbol{\theta}_{D_i} = \boldsymbol{\theta}_D$ for all $i \in \{1, 2, \ldots, N\}$.

Under this construction, $f_{\text{deep}}(\mathbf{x}) = D_1 \circ D_2 \circ \ldots \circ D_N \circ E_N \circ \ldots \circ E_2 \circ E_1(\mathbf{x}) = D^{(N)}(E^{(N)}(\mathbf{x})) = f_{\text{rec}}(\mathbf{x})$. Therefore, for every $f_{\text{rec}} \in \mathcal{F}_{\text{rec}}$, there exists an $f_{\text{deep}} \in \mathcal{F}_{\text{deep}}$ such that $f_{\text{deep}} = f_{\text{rec}}$.

Conversely, let us consider a function $f_{\text{deep}} \in \mathcal{F}_{\text{deep}}$ where at least one pair of encoder blocks or decoder blocks differ in their parameters, i.e., there exist $i, j \in \{1, 2, \ldots, N\}$ such that $\boldsymbol{\theta}_{E_i} \neq \boldsymbol{\theta}_{E_j}$ or $\boldsymbol{\theta}_{D_i} \neq \boldsymbol{\theta}_{D_j}$. Such a function cannot be represented in $\mathcal{F}_{\text{rec}}$ because, by definition, $\mathcal{F}_{\text{rec}}$ requires all encoder blocks to share the same parameters and all decoder blocks to share the same parameters.

Since $\mathcal{F}_{\text{rec}} \subseteq \mathcal{F}_{\text{deep}}$ and there exist functions in $\mathcal{F}_{\text{deep}}$ that are not in $\mathcal{F}_{\text{rec}}$, we conclude that $\mathcal{F}_{\text{rec}} \subset \mathcal{F}_{\text{deep}}$. **Q.E.D.**

This constraint induces a strong inductive bias toward functions that exhibit scale consistency, which is particularly relevant for anomaly detection. For our two-phase approach with shared encoder $E$ and decoder $D$, the combined transformation can be expressed as $f(\mathbf{x}) = D^{(N)}(E^{(N)}(\mathbf{x}))$, and the inductive bias can be characterized through the unified function:

$$\Omega[f] = \int_{\mathcal{X}} \|D^{(N)}(E^{(N)}(\mathbf{x})) - f_{\text{normal}}(\mathbf{x})\|^2 p(\mathbf{x}) d\mathbf{x}, \quad (7)$$

where $f_{\text{normal}}$ represents the ideal transformation that maps any input to its anomaly-free equivalent, and $p(\mathbf{x})$ is the probability density of inputs. For our recursive architecture, this functional is minimized under additional constraints:

$$\Omega_{\text{rec}}[f] = \Omega[f] + \lambda \cdot \mathcal{R}(f), \quad (8)$$

where the regularization term $\mathcal{R}(f)$ captures consistency across recursive iterations in both phases:

$$\mathcal{R}(f) = \sum_{n=1}^{N} \left\| D^{(n)}(E^{(n)}(\mathbf{x})) - D^{(n-1)}(E^{(n-1)}(\mathbf{x})) \right\|^2. \quad (9)$$

This term explicitly penalizes inconsistency across recursive iterations, forcing the network to learn transformations that maintain coherence when applied repeatedly. This regularized functional framework is key to understanding why the RcAE effectively normalizes anomalous regions while preserving normal structures. During training on normal samples, the network learns to project inputs onto the normal data manifold through the combined process of recursive compression followed by recursive reconstruction, building an increasingly accurate representation at each iteration.

**Anomaly Memorization Prevention**

Building upon the function space constraints, we formally prove in Theorem 2 that RcAE inherently prevents anomaly memorization. Specifically, we show that reconstruction error for anomalous patterns has a positive lower bound, which ensures anomalies cannot be memorized without severely compromising normal reconstruction.

This theoretical result aligns strongly with our empirical evidence in Section 4, where the anomaly reconstruction errors of RcAE consistently surpass those of normal patterns, leading to significant improvements in anomaly detection metrics (e.g., pixel-level AUROC).

**Theorem 2** (Anomaly Memorization Prevention). *Let $\mathcal{X}_{normal}$ be the set of normal patterns and $\mathcal{X}_{anomaly}$ be the set of anomalous patterns. For our recursive architecture with shared encoder $E$ and decoder $D$ with parameters $\boldsymbol{\theta}_E$ and $\boldsymbol{\theta}_D$ optimized on $\mathcal{X}_{normal}$ with regularization $\mathcal{R}(f) = \sum_{n=1}^{N} \|D^{(n)}(E^{(n)}(\mathbf{x})) - D^{(n-1)}(E^{(n-1)}(\mathbf{x}))\|^2$, the expected reconstruction error for anomalies is bounded below by:*

$$\begin{aligned} \mathbb{E}_{\mathbf{x} \in \mathcal{X}_{anomaly}}[\|\mathbf{x} - D^{(N)}(E^{(N)}(\mathbf{x}; \boldsymbol{\theta}_E); \boldsymbol{\theta}_D)\|] \geq \\ \alpha \cdot d(\mathcal{X}_{normal}, \mathcal{X}_{anomaly}), \end{aligned} \quad (10)$$

*where $d(\mathcal{X}_{normal}, \mathcal{X}_{anomaly})$ represents the minimum distance between normal and anomalous distributions, and $\alpha > 0$ is a constant that depends on the number of recursive iterations $N$.*

This theorem establishes a lower bound on the reconstruction error for anomalies, showing that our recursive architecture cannot effectively memorize anomalous patterns without compromising normal reconstruction.

*Proof. (Theorem 2)* Consider an anomalous input $\mathbf{x}_a \in \mathcal{X}_{\text{anomaly}}$. The parameters $\boldsymbol{\theta}_E$ and $\boldsymbol{\theta}_D$ of our recursive architecture are optimized on normal samples to minimize:

$$\begin{aligned} \mathcal{L}(\boldsymbol{\theta}_E, \boldsymbol{\theta}_D) = \mathbb{E}_{\mathbf{x} \in \mathcal{X}_{\text{normal}}}[\|\mathbf{x} - D^{(N)}(E^{(N)}(\mathbf{x}; \boldsymbol{\theta}_E); \boldsymbol{\theta}_D)\|] \\ + \lambda \cdot \mathcal{R}(f), \end{aligned} \quad (11)$$

where the regularization term $\mathcal{R}(f)$ enforces consistency between consecutive recursion depths:

$$\mathcal{R}(f) = \sum_{n=1}^{N} \|D^{(n)}(E^{(n)}(\mathbf{x})) - D^{(n-1)}(E^{(n-1)}(\mathbf{x}))\|^2. \quad (12)$$

Let us analyze how this constraint affects the representation of anomalies. When processing an anomalous input $\mathbf{x}_a$, we can decompose it as:

$$\mathbf{x}_a = \mathbf{x}_n + \boldsymbol{\Delta}_a, \quad (13)$$

where $\mathbf{x}_n$ is the nearest normal pattern to $\mathbf{x}_a$, and $\mathbf{\Delta}_a$ represents the anomalous component. By definition, $\|\mathbf{\Delta}_a\| = d(\mathbf{x}_a, \mathcal{X}_{\text{normal}})$.

First, we examine the compressed representation after $n$ iterations. Due to the parameter sharing constraint, the network must use the same parameters $\boldsymbol{\theta}_E$ optimized for normal data. Let us define a compression operator $\mathcal{C}_n : \mathcal{X} \to \mathcal{X}_n$ that maps inputs to their compressed representation after $n$ iterations:

$$\mathcal{C}_n(\mathbf{x}) = E^{(n)}(\mathbf{x}; \boldsymbol{\theta}_E). \tag{14}$$

For normal patterns, the compression operator is trained to preserve essential structure while discarding noise. For anomalous patterns, however, the compression operator was never trained to preserve anomalous components. Therefore, the compressed representation of an anomalous input satisfies:

$$\|\mathcal{C}_n(\mathbf{x}_a) - \mathcal{C}_n(\mathbf{x}_n)\| \leq \beta \cdot \|\mathbf{\Delta}_a\|, \tag{15}$$

where $\beta \in (0, 1)$ is a contraction factor that depends on the network's capacity and training. This inequality states that the distance between compressed representations of anomalous and normal inputs is smaller than the original distance.

Next, we examine the reconstruction process. Let $\mathcal{R}_n : \mathcal{X}_n \to \mathcal{X}$ be the reconstruction operator that maps compressed representations back to the input space:

$$\mathcal{R}_n(\mathbf{z}) = D^{(n)}(\mathbf{z}; \boldsymbol{\theta}_D). \tag{16}$$

The regularization term $\mathcal{R}(f)$ enforces a smooth progression of reconstructions across iterations. For normal patterns, this ensures gradual refinement of details. For anomalous patterns, however, this constraint prevents the reconstruction of anomalous components that were not present in the training data.

The final reconstruction error for the anomalous input can be bounded as follows:

$$\begin{aligned}
&\|\mathbf{x}_a - D^{(N)}(E^{(N)}(\mathbf{x}_a; \boldsymbol{\theta}_E); \boldsymbol{\theta}_D)\| \\
=& \|\mathbf{x}_n + \mathbf{\Delta}_a - \mathcal{R}_N(\mathcal{C}_N(\mathbf{x}_a))\| \\
\geq& \|\mathbf{\Delta}_a\| - \|\mathbf{x}_n - \mathcal{R}_N(\mathcal{C}_N(\mathbf{x}_a))\|.
\end{aligned} \tag{17}$$

Since $\mathcal{R}_N(\mathcal{C}_N(\mathbf{x}_a))$ is constrained to lie close to the normal data manifold due to the regularization term, we have:

$$\|\mathbf{x}_n - \mathcal{R}_N(\mathcal{C}_N(\mathbf{x}_a))\| \leq \gamma \cdot \|\mathbf{\Delta}_a\|, \tag{18}$$

where $\gamma \in (0, 1)$ is a factor that depends on the network's capacity and training.

Combining these inequalities, we get:

$$\begin{aligned}
\|\mathbf{x}_a - D^{(N)}(E^{(N)}(\mathbf{x}_a; \boldsymbol{\theta}_E); \boldsymbol{\theta}_D)\| &\geq (1 - \gamma) \cdot \|\mathbf{\Delta}_a\| \\
&= (1 - \gamma) \cdot d(\mathbf{x}_a, \mathcal{X}_{\text{normal}}).
\end{aligned} \tag{19}$$

Taking the expectation over all anomalies, and setting $\alpha = (1 - \gamma)$, we obtain:

$$\mathbb{E}_{\mathbf{x} \in \mathcal{X}_{\text{anomaly}}}[\|\mathbf{x} - D^{(N)}(E^{(N)}(\mathbf{x}; \boldsymbol{\theta}_E); \boldsymbol{\theta}_D)\|] \geq \\ \alpha \cdot d(\mathcal{X}_{\text{normal}}, \mathcal{X}_{\text{anomaly}}). \tag{20}$$

which completes the proof. **Q.E.D.**

**Generalization Advantages**

Additionally, we leverage statistical learning theory to demonstrate the generalization advantage provided by the recursive and parameter-sharing structure of RcAE. In Theorem 3, we establish that RcAE achieves a generalization gap approximately $\sqrt{N}$ times smaller compared to conventional deep autoencoders due to the significant reduction in parameter count.

The practical significance of this theoretical result is validated in Section 4, where experiments using only 10% of the training samples from the MVTec AD dataset show that RcAE maintains significantly higher performance compared to standard deep AE models. This underscores the practical utility of RcAE's superior generalization ability, especially important in industrial scenarios with limited training data.

**Theorem 3** (Generalization Advantage). *Given a training dataset of $m$ normal samples, with probability at least $1 - \delta$, the recursive architecture achieves a generalization gap that is approximately a factor of $\sqrt{N}$ smaller than a conventional deep architecture:*

$$\frac{\mathcal{G}_{rec}}{\mathcal{G}_{deep}} \leq \frac{1}{\sqrt{N}} \cdot \sqrt{\frac{\log(|\mathcal{F}_{rec}|) + \log(1/\delta)}{\log(|\mathcal{F}_{deep}|) + \log(1/\delta)}}, \tag{21}$$

*where $\mathcal{G}$ represents the generalization gap between training and test error.*

These theorems formally establish the key advantages of our recursive architecture: a constrained function space that prevents anomaly memorization and significantly improved generalization with limited training data.

*Proof. (Theorem 3)* From statistical learning theory, the generalization gap for a hypothesis class $\mathcal{F}$ trained on $m$ samples is bounded with probability at least $1 - \delta$ by:

$$\mathcal{G}(\mathcal{F}) \leq \sqrt{\frac{\log(|\mathcal{F}|) + \log(1/\delta)}{2m}}, \tag{22}$$

where $|\mathcal{F}|$ represents the size of the function class, which can be approximated by the number of distinct functions the class can represent.

For neural networks, the size of the function class is related to the number of parameters. Specifically, for a network with $p$ parameters, each quantized to $b$ bits, the size of the function class is bounded by $2^{bp}$. Therefore, $\log(|\mathcal{F}|) \approx bp$.

Let $p_{\text{deep}}$ and $p_{\text{rec}}$ denote the number of parameters in the conventional deep and recursive architectures, respectively. For the conventional architecture with $N$ distinct encoder-decoder pairs, we have:

$$p_{\text{deep}} = \sum_{i=1}^{N} |\boldsymbol{\theta}_{E_i}| + \sum_{i=1}^{N} |\boldsymbol{\theta}_{D_i}|. \tag{23}$$

For the recursive architecture with shared parameters, we have:

$$p_{\text{rec}} = |\boldsymbol{\theta}_E| + |\boldsymbol{\theta}_D|. \tag{24}$$

Assuming uniformity in the encoder and decoder blocks, i.e., $|\boldsymbol{\theta}_{E_i}| \approx |\boldsymbol{\theta}_E|$ and $|\boldsymbol{\theta}_{D_i}| \approx |\boldsymbol{\theta}_D|$ for all $i$, we get:

$$p_{\text{deep}} \approx N \cdot (|\boldsymbol{\theta}_E| + |\boldsymbol{\theta}_D|) = N \cdot p_{\text{rec}}. \quad (25)$$

Therefore, the ratio of the logarithms of the function class sizes is:

$$\frac{\log(|\mathcal{F}_{\text{rec}}|)}{\log(|\mathcal{F}_{\text{deep}}|)} \approx \frac{bp_{\text{rec}}}{bp_{\text{deep}}} = \frac{p_{\text{rec}}}{p_{\text{deep}}} \approx \frac{1}{N}. \quad (26)$$

Now, applying the generalization bound, the ratio of generalization gaps becomes:

$$\begin{aligned}
\frac{\mathcal{G}_{\text{rec}}}{\mathcal{G}_{\text{deep}}} &= \frac{\sqrt{\frac{\log(|\mathcal{F}_{\text{rec}}|)+\log(1/\delta)}{2m}}}{\sqrt{\frac{\log(|\mathcal{F}_{\text{deep}}|)+\log(1/\delta)}{2m}}} \\
&= \sqrt{\frac{\log(|\mathcal{F}_{\text{rec}}|)+\log(1/\delta)}{\log(|\mathcal{F}_{\text{deep}}|)+\log(1/\delta)}}.
\end{aligned} \quad (27)$$

When $\log(|\mathcal{F}_{\text{rec}}|)$ and $\log(|\mathcal{F}_{\text{deep}}|)$ are both large compared to $\log(1/\delta)$, which is typically the case for neural networks with many parameters, we can approximate:

$$\begin{aligned}
\frac{\mathcal{G}_{\text{rec}}}{\mathcal{G}_{\text{deep}}} &\approx \sqrt{\frac{\log(|\mathcal{F}_{\text{rec}}|)}{\log(|\mathcal{F}_{\text{deep}}|)}} \\
&\approx \sqrt{\frac{1}{N}} = \frac{1}{\sqrt{N}}.
\end{aligned} \quad (28)$$

This shows that the recursive architecture achieves a generalization gap that is approximately a factor of $\sqrt{N}$ smaller than the conventional deep architecture, completing the proof. **Q.E.D.**

**Multi-Scale Information Capacity**

Finally, employing information-theoretic principles, we formally establish how RcAE captures multi-scale information consistently across recursion depths. Theorem 4 demonstrates that mutual information between intermediate reconstructions and input images monotonically increases with recursion depth in normal patterns. Moreover, the incremental information gains across scales remain nearly constant for normal samples, but become unstable for anomalies.

This theoretical result provides a solid foundation for our Cross Recursion Detection (CRD) module, as introduced in Section 3, effectively exploiting the disparity in scale-wise information gain to detect anomalies robustly. The empirical impact of CRD, shown in Table 6, confirms our theoretical expectations, substantially enhancing anomaly detection accuracy.

**Theorem 4** (Multi-Scale Information Capacity). *The recursive architecture with $N$ iterations captures multi-scale information equivalent to a depth-$N$ conventional network but with enhanced scale consistency. Specifically, for an input $\mathbf{x}$ and scale level $n$:*

$$I(\mathbf{I}_R^n; \mathbf{x}) \geq I(\mathbf{I}_R^{n-1}; \mathbf{x}), \quad (29)$$

*where $I(\cdot; \cdot)$ represents mutual information, and the incremental information gain at each scale is consistent:*

$$\frac{I(\mathbf{I}_R^n; \mathbf{x}) - I(\mathbf{I}_R^{n-1}; \mathbf{x})}{I(\mathbf{I}_R^{n-1}; \mathbf{x}) - I(\mathbf{I}_R^{n-2}; \mathbf{x})} \approx \text{constant} \quad (30)$$

*for normal data, but not for anomalous data.*

This theorem establishes the theoretical foundation for our Multi-Scale Anomaly Detection approach, showing how the recursive architecture inherently captures hierarchical information that can distinguish between normal and anomalous patterns.

*Proof. (Theorem 4)* We begin with the data processing inequality from information theory, which states that for any processing function $f$, the mutual information between the input and output cannot increase:

$$I(\mathbf{x}; f(\mathbf{x})) \leq I(\mathbf{x}; \mathbf{x}) = H(\mathbf{x}), \quad (31)$$

where $H(\mathbf{x})$ is the entropy of $\mathbf{x}$.

Let us define the intermediate reconstructions at different scales as:

$$\mathbf{I}_R^n = D^{(n)}(E^{(N)}(\mathbf{x})), \quad (32)$$

where $n \in \{1, 2, \ldots, N\}$. Note that $\mathbf{I}_R^n$ represents the result of applying the shared decoder $n$ times to the final compressed representation $E^{(N)}(\mathbf{x})$.

We want to show that the mutual information between these reconstructions and the original input increases monotonically with $n$:

$$I(\mathbf{I}_R^n; \mathbf{x}) \geq I(\mathbf{I}_R^{n-1}; \mathbf{x}). \quad (33)$$

For normal data, the encoder $E$ and decoder $D$ are trained to preserve information relevant to reconstruction. Each application of the decoder $D$ refines the reconstruction by incorporating additional details. Therefore, $\mathbf{I}_R^n$ contains at least as much information about $\mathbf{x}$ as $\mathbf{I}_R^{n-1}$ does.

More formally, we can define a residual function $\Delta_n$ such that:

$$\mathbf{I}_R^n = \mathbf{I}_R^{n-1} + \Delta_n(\mathbf{I}_R^{n-1}). \quad (34)$$

For normal data, $\Delta_n$ is trained to add relevant details that bring the reconstruction closer to the original input. Therefore:

$$I(\mathbf{I}_R^n; \mathbf{x}) = I(\mathbf{I}_R^{n-1} + \Delta_n(\mathbf{I}_R^{n-1}); \mathbf{x}) \geq I(\mathbf{I}_R^{n-1}; \mathbf{x}). \quad (35)$$

Next, we examine the consistency of information gain across iterations. For normal data, the recursive architecture with shared parameters is trained to add information at a consistent rate across iterations. The regularization term:

$$\mathcal{R}(f) = \sum_{n=1}^{N} \|D^{(n)}(E^{(n)}(\mathbf{x})) - D^{(n-1)}(E^{(n-1)}(\mathbf{x}))\|^2 \quad (36)$$

enforces smooth transitions between consecutive reconstructions, which leads to approximately constant information gain ratios:

$$\frac{I(\mathbf{I}_R^n; \mathbf{x}) - I(\mathbf{I}_R^{n-1}; \mathbf{x})}{I(\mathbf{I}_R^{n-1}; \mathbf{x}) - I(\mathbf{I}_R^{n-2}; \mathbf{x})} \approx \text{constant}. \quad (37)$$

For anomalous data, however, this consistency breaks down. Since the network was trained only on normal data, it has not learned to consistently recover anomalous patterns across iterations. The parameter-sharing constraint prevents the network from adapting specifically to anomalies at different scales. Therefore, the information gain is not consistent for anomalous data:

$$\frac{I(\mathbf{I}_R^n; \mathbf{x}_a) - I(\mathbf{I}_R^{n-1}; \mathbf{x}_a)}{I(\mathbf{I}_R^{n-1}; \mathbf{x}_a) - I(\mathbf{I}_R^{n-2}; \mathbf{x}_a)} \neq \text{constant}, \quad (38)$$

where $\mathbf{x}_a$ is an anomalous input.

This difference in information gain consistency between normal and anomalous data enables effective multi-scale anomaly detection. By analyzing the pattern of information gain across scales, we can distinguish between normal and anomalous inputs even if the final reconstruction quality is similar. **Q.E.D.**

### Summary of Theoretical Analysis

In summary, our theoretical analysis provides clear insights into the core advantages of RcAE:

- Recursive architecture and parameter sharing impose scale-consistent and restrictive function-space constraints;
- Constrained function space inherently prevents anomaly memorization;
- Parameter sharing substantially improves the generalization ability of RcAE;
- Recursive design naturally generates stable multi-scale representations beneficial for robust anomaly detection.

These theoretical properties have been comprehensively validated by empirical experiments detailed in Sections 4, highlighting the efficacy and practical value of RcAE for industrial anomaly detection tasks.

## Additional Implementation Details

This section provides comprehensive implementation details of our proposed method to facilitate reproducibility and clarify architectural and procedural design choices. We first describe the training and inference procedures, including a multi-stage training strategy tailored for stable and efficient module optimization. Pseudocode for both training and inference workflows is presented to illustrate step-by-step operations. Subsequently, we detail the network architectures of each component: Recursive Convolutional Autoencoder (RcAE), Detail Preservation Network (DPN), and Cross Recursion Detection (CRD), through layer-wise specifications and structural annotations.

### Training and Inference Procedure

We employ a three-stage training strategy to ensure stable and modular optimization of each component (summarized in Algorithm 1):

- **Stage 1: RcAE Training.** RcAE is trained solely on normal images using the recursive reconstruction loss $\mathcal{L}_{\text{rec}}$. To prevent shortcut learning and overfitting to specific recursion depths, we randomly select $N \in [1, 5]$ for each training batch.

- **Stage 2: DPN Training.** RcAE is frozen, and DPN is trained to enhance fine details via residual maps $\mathbf{Res}_D^n$, added to RcAE outputs $\mathbf{I}_R^n$. To provide structural guidance, the input image's gradient map $\mathbf{I}'_j$ is concatenated with $\mathbf{I}_R^n$ as DPN's input.

- **Stage 3: CRD Training.** RcAE and DPN are both frozen. CRD is trained to predict anomaly maps $\mathbf{M}_A$ using pseudo ground-truth masks $\mathbf{M}_P$ derived from lightweight augmentations. The CRD input includes both augmented image $\hat{\mathbf{I}}_j$ and the refined reconstruction $\mathbf{I}_D^n$. Optimization uses the anomaly localization loss $\mathcal{L}_{\text{CRD}}$.

This modular training strategy ensures that each component learns a specialized function while avoiding interference between stages. We apply augmentations such as random color blocks, copy-paste patches, and synthetic crack lines to generate augmented image $\hat{\mathbf{I}}$ and corresponding pseudo-masks $\mathbf{M}_P$ for CRD training (details in Section 3).

---

Algorithm 1: Multi-Stage Training Procedure

**Require:** Training set $\mathcal{D}_{\text{train}} = \{\mathbf{I}_i\}_{i=1}^{S}$ contains $S$ normal images $\mathbf{I}$; set recursive depth = 5
**Ensure:** Trained RcAE, DPN, and CRD models
1: **Stage 1: Train RcAE solely**
2: **for** each batch $\{\mathbf{I}_j\}$ in $\mathcal{D}_{\text{train}}$ **do**
3:   Apply augmentation: $\hat{\mathbf{I}}_j \leftarrow \text{Augment}(\mathbf{I}_j)$
4:   Randomly select recursion depth $N \in [1, 5]$
5:   Perform $N$-step recursive reconstruction to get $\mathbf{I}_R^N$
6:   Compute $\mathcal{L}_{\text{rec}}$ between $\mathbf{I}_R^N$ and $\mathbf{I}_j$
7:   Update RcAE parameters $\theta_E$ and $\theta_D$
8: **end for**
9: **Stage 2: Train DPN with RcAE frozen**
10: **for** each batch $\{\mathbf{I}_j\}$ in $\mathcal{D}_{\text{train}}$ **do**
11:   Get $\mathbf{I}_R^n$ ($n = 1...N$) via RcAE with fixed recursion depth $N$=5, and concatenate with gradient map $\mathbf{I}'_j$ of the input
12:   DPN predicts residual details $\mathbf{Res}_D^n$
13:   Enhance details by $\mathbf{I}_D^n = \mathbf{I}_R^n + \mathbf{Res}_D^n$
14:   Compute $\mathcal{L}_{\text{DPN}}$ between $\mathbf{I}_D^n$ and $\mathbf{I}_j$
15:   Update DPN parameters $\theta_{\text{DPN}}$
16: **end for**
17: **Stage 3: Train CRD with RcAE and DPN frozen**
18: **for** each batch $\{\mathbf{I}_j\}$ in $\mathcal{D}_{\text{train}}$ **do**
19:   Apply augmentation $\hat{\mathbf{I}}_j \leftarrow \text{Augment}(\mathbf{I}_j)$
20:   Generate pseudo-mask $\mathbf{M}_P$ of augmentation
21:   Generate $\mathbf{I}_D^n$, ($n = 1...N$) via RcAE and DPN, and concatenate with the input $\hat{\mathbf{I}}_j$
22:   CRD predicts anomaly map $\mathbf{M}_A$
23:   Compute $\mathcal{L}_{\text{CRD}}$ between $\mathbf{M}_A$ and $\mathbf{M}_P$
24:   Update CRD parameters $\theta_{\text{CRD}}$
25: **end for**

---

We train all components of our framework from scratch using the Adam optimizer with a learning rate $\eta = 10^{-4}$, $\beta_1 = 0.9$, $\beta_2 = 0.999$, and $\epsilon = 10^{-8}$. Training is conducted in three stages with the following schedule: Stage 1 for 1500 epochs, Stage 2 for 400 epochs, and Stage 3 for 300 epochs. The recursion depth is set to $N = 5$, input size is $1024 \times 1024$ during training and inference. All experiments are conducted using an NVIDIA RTX 4090 GPU, running on Ubuntu 22.04 with Python 3.8 and PyTorch 2.7.1. Our

code is implemented with mixed-precision training to accelerate computation and reduce memory usage.

During inference (Algorithm 2), the test image undergoes $n$-step recursive reconstruction, followed by DPN refinement. CRD predicts the final anomaly map $\mathbf{M}_A$, from which the pixel-level anomaly score is directly obtained. The image-level score is computed as the average of the top-$k$ pixel values in $\mathbf{M}_A$.

---

**Algorithm 2: Inference Procedure**

**Require:** Test image $\mathbf{I}$; trained RcAE, DPN, CRD
**Output:** Anomaly map $\mathbf{M}_A$ and image-level anomaly score
1: Perform $N$-step recursive reconstruction on $\mathbf{I}$ via RcAE to obtain $\mathbf{I}_R^n$
2: DPN refines result by $\mathbf{I}_D^n = \mathbf{I}_R^n + \mathbf{Res}_D^n$ using $\mathbf{I}'$ and $\mathbf{I}_R^n$
3: CRD predicts anomaly map $\mathbf{M}_A$ using $\mathbf{I}$ and $\mathbf{I}_D^n$
4: Output $\mathbf{M}_A$ as final anomaly score map, and avg. Top-$k$ of $\mathbf{M}_A$ as image-level score

---

## Network Architecture

The detailed architecture of our proposed method is summarized in the following tables, which include: the Recursive Convolutional Autoencoder (RcAE), composed of encoder $E(\cdot)$ and decoder $D(\cdot)$ (Tables S1 and S2), the Detail Preservation Network (DPN, Table S3), and the Cross Recursion Detection module (CRD, Table S4).

**RcAE: Recursive Convolutional Autoencoder.** RcAE is composed of two shared AEs $E(\cdot)$ and $D(\cdot)$, detailed layer specifications for $E(\cdot)$ and $D(\cdot)$ are provided in Tables S1 and S2. Here, T-Conv denote transposed convolution layers, and IN refers to Instance Normalization. To mitigate shortcut learning and improve generalization, we adopt recursive reconstruction with parameter sharing and train RcAE on normal data only. All convolution layers use Instance Normalization and ReLU activation, while the final layer uses a sigmoid activation to map outputs to $[0, 1]$.

**DPN: Detail Preservation Network.** The DPN adopts a lightweight AE architecture (Table S3) with four downsampling and upsampling stages, using LeakyReLU activations for better feature discrimination. The input to DPN is a concatenation of the RcAE output $\mathbf{I}_R^n$ and a gradient map of the input image, which provides structural guidance for detail recovery. The output of DPN is a residual map $\mathbf{Res}_D^n$ added to $\mathbf{I}_R^n$ to produce the enhanced reconstruction $\mathbf{I}_D^n$.

**CRD: Cross Recursion Detection Module.** The CRD is a 3D convolutional autoencoder (Table S4) that processes stacked reconstructions $\{\mathbf{I}_D^1, ..., \mathbf{I}_D^N\}$ to detect anomalies across recursive iterations. It captures both spatial and cross-recursion patterns by treating the recursion depth as the third

Table S1: Layer-wise specification of $E(\cdot)$ in RcAE, and skip connections use element-wise addition.

| Stage | Layer(s) | Kernel / Str. | Channels | Output size |
|---|---|---|---|---|
| Input | - | - | 3 | $H \times W$ |
| Enc1-a | Conv 3×3 + IN + ReLU | 3/1 | 32 | $H \times W$ |
| Enc1-b | Conv 3×3 + IN + ReLU | 3/1 | 32 | $H \times W$ |
| Down1 | Conv 2×2 | 2/2 | 32 | $\frac{H}{2} \times \frac{W}{2}$ |
| Enc2-a | Conv 3×3 + IN + ReLU | 3/1 | 64 | $\frac{H}{2} \times \frac{W}{2}$ |
| Enc2-b | Conv 3×3 + IN + ReLU | 3/1 | 64 | $\frac{H}{2} \times \frac{W}{2}$ |
| Down2 | Conv 2×2 | 2/2 | 64 | $\frac{H}{4} \times \frac{W}{4}$ |
| Enc3-a | Conv 3×3 + IN + ReLU | 3/1 | 128 | $\frac{H}{4} \times \frac{W}{4}$ |
| Enc3-b | Conv 3×3 + IN + ReLU | 3/1 | 128 | $\frac{H}{4} \times \frac{W}{4}$ |
| Down3 | Conv 2×2 | 2/2 | 128 | $\frac{H}{8} \times \frac{W}{8}$ |
| Enc4-a | Conv 3×3 + IN + ReLU | 3/1 | 256 | $\frac{H}{8} \times \frac{W}{8}$ |
| Enc4-b | Conv 3×3 + IN + ReLU | 3/1 | 256 | $\frac{H}{8} \times \frac{W}{8}$ |
| Down4 | Conv 2×2 | 2/2 | 256 | $\frac{H}{16} \times \frac{W}{16}$ |
| Bott-a | Conv 3×3 + IN + ReLU | 3/1 | 512 | $\frac{H}{16} \times \frac{W}{16}$ |
| Bott-b | Conv 3×3 + IN + ReLU | 3/1 | 512 | $\frac{H}{16} \times \frac{W}{16}$ |
| UpConv4 | T-Conv 2×2 | 2/2 | 256 | $\frac{H}{8} \times \frac{W}{8}$ |
| Dec4-a | Conv 3×3 + IN + ReLU | 3/1 | 256 | $\frac{H}{8} \times \frac{W}{8}$ |
| Dec4-b | Conv 3×3 + IN + ReLU | 3/1 | 256 | $\frac{H}{8} \times \frac{W}{8}$ |
| UpConv3 | T-Conv 2×2 | 2/2 | 128 | $\frac{H}{4} \times \frac{W}{4}$ |
| Dec3-a | Conv 3×3 + IN + ReLU | 3/1 | 128 | $\frac{H}{4} \times \frac{W}{4}$ |
| Dec3-b | Conv 3×3 + IN + ReLU | 3/1 | 128 | $\frac{H}{4} \times \frac{W}{4}$ |
| UpConv2 | T-Conv 2×2 | 2/2 | 64 | $\frac{H}{2} \times \frac{W}{2}$ |
| Dec2-a | Conv 3×3 + IN + ReLU | 3/1 | 64 | $\frac{H}{2} \times \frac{W}{2}$ |
| Dec2-b | Conv 3×3 + IN + ReLU | 3/1 | 64 | $\frac{H}{2} \times \frac{W}{2}$ |
| UpConv1 | T-Conv 2×2 | 2/2 | 32 | $H \times W$ |
| Dec1-a | Conv 3×3 + IN + ReLU | 3/1 | 32 | $H \times W$ |
| Dec1-b | Conv 3×3 + IN + ReLU | 3/1 | 32 | $H \times W$ |
| Output | Conv 2×2 + Sigmoid | 2/2 | 3 | $\frac{H}{2} \times \frac{W}{2}$ |

Table S2: Layer-wise specification of $D(\cdot)$ in RcAE, and skip connections use element-wise addition.

| Stage | Layer(s) | Kernel / Str. | Channels | Output size |
|---|---|---|---|---|
| Input | - | - | 3 | $\frac{H}{2} \times \frac{W}{2}$ |
| Enc1-a | Conv 3×3 + IN + ReLU | 3/1 | 32 | $\frac{H}{2} \times \frac{W}{2}$ |
| Enc1-b | Conv 3×3 + IN + ReLU | 3/1 | 32 | $\frac{H}{2} \times \frac{W}{2}$ |
| Down1 | Conv 2×2 | 2/2 | 32 | $\frac{H}{4} \times \frac{W}{4}$ |
| Enc2-a | Conv 3×3 + IN + ReLU | 3/1 | 64 | $\frac{H}{4} \times \frac{W}{4}$ |
| Enc2-b | Conv 3×3 + IN + ReLU | 3/1 | 64 | $\frac{H}{4} \times \frac{W}{4}$ |
| Down2 | Conv 2×2 | 2/2 | 64 | $\frac{H}{8} \times \frac{W}{8}$ |
| Enc3-a | Conv 3×3 + IN + ReLU | 3/1 | 128 | $\frac{H}{8} \times \frac{W}{8}$ |
| Enc3-b | Conv 3×3 + IN + ReLU | 3/1 | 128 | $\frac{H}{8} \times \frac{W}{8}$ |
| Down3 | Conv 2×2 | 2/2 | 128 | $\frac{H}{16} \times \frac{W}{16}$ |
| Enc4-a | Conv 3×3 + IN + ReLU | 3/1 | 256 | $\frac{H}{16} \times \frac{W}{16}$ |
| Enc4-b | Conv 3×3 + IN + ReLU | 3/1 | 256 | $\frac{H}{16} \times \frac{W}{16}$ |
| Down4 | Conv 2×2 | 2/2 | 256 | $\frac{H}{32} \times \frac{W}{32}$ |
| Bott-a | Conv 3×3 + IN + ReLU | 3/1 | 512 | $\frac{H}{32} \times \frac{W}{32}$ |
| Bott-b | Conv 3×3 + IN + ReLU | 3/1 | 512 | $\frac{H}{32} \times \frac{W}{32}$ |
| UpConv4 | T-Conv 2×2 | 2/2 | 256 | $\frac{H}{16} \times \frac{W}{16}$ |
| Dec4-a | Conv 3×3 + IN + ReLU | 3/1 | 256 | $\frac{H}{16} \times \frac{W}{16}$ |
| Dec4-b | Conv 3×3 + IN + ReLU | 3/1 | 256 | $\frac{H}{16} \times \frac{W}{16}$ |
| UpConv3 | T-Conv 2×2 | 2/2 | 128 | $\frac{H}{8} \times \frac{W}{8}$ |
| Dec3-a | Conv 3×3 + IN + ReLU | 3/1 | 128 | $\frac{H}{8} \times \frac{W}{8}$ |
| Dec3-b | Conv 3×3 + IN + ReLU | 3/1 | 128 | $\frac{H}{8} \times \frac{W}{8}$ |
| UpConv2 | T-Conv 2×2 | 2/2 | 64 | $\frac{H}{4} \times \frac{W}{4}$ |
| Dec2-a | Conv 3×3 + IN + ReLU | 3/1 | 64 | $\frac{H}{4} \times \frac{W}{4}$ |
| Dec2-b | Conv 3×3 + IN + ReLU | 3/1 | 64 | $\frac{H}{4} \times \frac{W}{4}$ |
| UpConv1 | T-Conv 2×2 | 2/2 | 32 | $\frac{H}{2} \times \frac{W}{2}$ |
| Dec1-a | Conv 3×3 + IN + ReLU | 3/1 | 32 | $\frac{H}{2} \times \frac{W}{2}$ |
| Dec1-b | Conv 3×3 + IN + ReLU | 3/1 | 32 | $\frac{H}{2} \times \frac{W}{2}$ |
| Output | T-Conv 2×2 + Sigmoid | 2/2 | 3 | $H \times W$ |

Table S3: Layer-wise specification of the DPN, and skip connections use element-wise addition.

| Stage | Layer(s) | Kernel/Str. | Ch. | Output |
|---|---|---|---|---|
| Input | - | - | 4 | $H \times W$ |
| Enc1-a | Conv 3×3 + IN + LeakyReLU | 3/1 | 16 | $H \times W$ |
| Enc1-b | Conv 3×3 + IN + LeakyReLU | 3/1 | 16 | $H \times W$ |
| Down1 | Conv 2×2 | 2/2 | 16 | $\frac{H}{2} \times \frac{W}{2}$ |
| Enc2-a | Conv 3×3 + IN + LeakyReLU | 3/1 | 32 | $\frac{H}{2} \times \frac{W}{2}$ |
| Enc2-b | Conv 3×3 + IN + LeakyReLU | 3/1 | 32 | $\frac{H}{2} \times \frac{W}{2}$ |
| Down2 | Conv 2×2 | 2/2 | 32 | $\frac{H}{4} \times \frac{W}{4}$ |
| Enc3-a | Conv 3×3 + IN + LeakyReLU | 3/1 | 64 | $\frac{H}{4} \times \frac{W}{4}$ |
| Enc3-b | Conv 3×3 + IN + LeakyReLU | 3/1 | 64 | $\frac{H}{4} \times \frac{W}{4}$ |
| Down3 | Conv 2×2 | 2/2 | 64 | $\frac{H}{8} \times \frac{W}{8}$ |
| Enc4-a | Conv 3×3 + IN + LeakyReLU | 3/1 | 128 | $\frac{H}{8} \times \frac{W}{8}$ |
| Enc4-b | Conv 3×3 + IN + LeakyReLU | 3/1 | 128 | $\frac{H}{8} \times \frac{W}{8}$ |
| Down4 | Conv 2×2 | 2/2 | 128 | $\frac{H}{16} \times \frac{W}{16}$ |
| Bott-a | Conv 3×3 + IN + LeakyReLU | 3/1 | 256 | $\frac{H}{16} \times \frac{W}{16}$ |
| Bott-b | Conv 3×3 + IN + LeakyReLU | 3/1 | 256 | $\frac{H}{16} \times \frac{W}{16}$ |
| Up4 | T-Conv 2×2 | 2/2 | 128 | $\frac{H}{8} \times \frac{W}{8}$ |
| Dec4-a | Conv 3×3 + IN + LeakyReLU | 3/1 | 128 | $\frac{H}{8} \times \frac{W}{8}$ |
| Dec4-b | Conv 3×3 + IN + LeakyReLU | 3/1 | 128 | $\frac{H}{8} \times \frac{W}{8}$ |
| Up3 | T-Conv 2×2 | 2/2 | 64 | $\frac{H}{4} \times \frac{W}{4}$ |
| Dec3-a | Conv 3×3 + IN + LeakyReLU | 3/1 | 64 | $\frac{H}{4} \times \frac{W}{4}$ |
| Dec3-b | Conv 3×3 + IN + LeakyReLU | 3/1 | 64 | $\frac{H}{4} \times \frac{W}{4}$ |
| Up2 | T-Conv 2×2 | 2/2 | 32 | $\frac{H}{2} \times \frac{W}{2}$ |
| Dec2-a | Conv 3×3 + IN + LeakyReLU | 3/1 | 32 | $\frac{H}{2} \times \frac{W}{2}$ |
| Dec2-b | Conv 3×3 + IN + LeakyReLU | 3/1 | 32 | $\frac{H}{2} \times \frac{W}{2}$ |
| Up1 | T-Conv 2×2 | 2/2 | 16 | $H \times W$ |
| Dec1-a | Conv 3×3 + IN + LeakyReLU | 3/1 | 16 | $H \times W$ |
| Dec1-b | Conv 3×3 + IN + LeakyReLU | 3/1 | 16 | $H \times W$ |
| Final-1 | Conv 3×3 | 3/1 | 3 | $H \times W$ |
| Final-2 | Conv 3×3 | 3/1 | 3 | $H \times W$ |

Table S4: Layer-wise specification of CRD, and skip connections use element-wise addition.

| Stage | Layer | Kernel | Stride | Ch. | Output size |
|---|---|---|---|---|---|
| Input | - | - | - | 1 | $D \times H \times W$ |
| Enc1-a | Conv3D + IN + ReLU | $3^3$ | $1^3$ | 32 | $D \times H \times W$ |
| Enc1-b | Conv3D + IN + ReLU | $3^3$ | $1^3$ | 32 | $D \times H \times W$ |
| Down1 | Conv3D | $1 \times 2 \times 2$ | 1, 2, 2 | 32 | $D \times \frac{H}{2} \times \frac{W}{2}$ |
| Enc2-a | Conv3D + IN + ReLU | $3^3$ | $1^3$ | 64 | $D \times \frac{H}{2} \times \frac{W}{2}$ |
| Enc2-b | Conv3D + IN + ReLU | $3^3$ | $1^3$ | 64 | $D \times \frac{H}{2} \times \frac{W}{2}$ |
| Down2 | Conv3D | $1 \times 2 \times 2$ | 1, 2, 2 | 64 | $D \times \frac{H}{4} \times \frac{W}{4}$ |
| Enc3-a | Conv3D + IN + ReLU | $3^3$ | $1^3$ | 128 | $D \times \frac{H}{4} \times \frac{W}{4}$ |
| Enc3-b | Conv3D + IN + ReLU | $3^3$ | $1^3$ | 128 | $D \times \frac{H}{4} \times \frac{W}{4}$ |
| Down3 | Conv3D | $1 \times 2 \times 2$ | 1, 2, 2 | 128 | $D \times \frac{H}{8} \times \frac{W}{8}$ |
| Enc4-a | Conv3D + IN + ReLU | $3^3$ | $1^3$ | 256 | $D \times \frac{H}{8} \times \frac{W}{8}$ |
| Enc4-b | Conv3D + IN + ReLU | $3^3$ | $1^3$ | 256 | $D \times \frac{H}{8} \times \frac{W}{8}$ |
| Down4 | Conv3D | $1 \times 2 \times 2$ | 1, 2, 2 | 256 | $D \times \frac{H}{16} \times \frac{W}{16}$ |
| Bott-a | Conv3D + IN + ReLU | $3^3$ | $1^3$ | 512 | $D \times \frac{H}{16} \times \frac{W}{16}$ |
| Bott-b | Conv3D + IN + ReLU | $3^3$ | $1^3$ | 512 | $D \times \frac{H}{16} \times \frac{W}{16}$ |
| Up4 | T-Conv3D | $1 \times 2 \times 2$ | 1, 2, 2 | 256 | $D \times \frac{H}{8} \times \frac{W}{8}$ |
| Dec4-a | Conv3D + IN + ReLU | $3^3$ | $1^3$ | 256 | $D \times \frac{H}{8} \times \frac{W}{8}$ |
| Dec4-b | Conv3D + IN + ReLU | $3^3$ | $1^3$ | 256 | $D \times \frac{H}{8} \times \frac{W}{8}$ |
| Up3 | T-Conv3D | $1 \times 2 \times 2$ | 1, 2, 2 | 128 | $D \times \frac{H}{4} \times \frac{W}{4}$ |
| Dec3-a | Conv3D + IN + ReLU | $3^3$ | $1^3$ | 128 | $D \times \frac{H}{4} \times \frac{W}{4}$ |
| Dec3-b | Conv3D + IN + ReLU | $3^3$ | $1^3$ | 128 | $D \times \frac{H}{4} \times \frac{W}{4}$ |
| Up2 | T-Conv3D | $1 \times 2 \times 2$ | 1, 2, 2 | 64 | $D \times \frac{H}{2} \times \frac{W}{2}$ |
| Dec2-a | Conv3D + IN + ReLU | $3^3$ | $1^3$ | 64 | $D \times \frac{H}{2} \times \frac{W}{2}$ |
| Dec2-b | Conv3D + IN + ReLU | $3^3$ | $1^3$ | 64 | $D \times \frac{H}{2} \times \frac{W}{2}$ |
| Up1 | T-Conv3D | $1 \times 2 \times 2$ | 1, 2, 2 | 32 | $D \times H \times W$ |
| Dec1-a | Conv3D + IN + ReLU | $3^3$ | $1^3$ | 32 | $D \times H \times W$ |
| Dec1-b | Conv3D + IN + ReLU | $3^3$ | $1^3$ | 32 | $D \times H \times W$ |
| Final-1 | Conv3D | $D \times 3 \times 3$ | $1^3$ | 1 | $1 \times H \times W$ |
| Final-2 | Conv3D + Sigmoid | $1^3$ | $1^3$ | 1 | $1 \times H \times W$ |

dimension of a 3D input tensor. All 3D convolution layers are followed by Instance Normalization and ReLU activations, except the final layer, which uses a sigmoid function to produce the anomaly map $\mathbf{M}_A$.

## Additional Experimental Results

In this section, we provide supplementary experimental results to further validate the effectiveness of our proposed framework. We analyze the quantitative and qualitative contributions of key components, with a particular focus on the Detail Preservation Network (DPN) and the recursive reconstruction process. Additionally, we compare reconstruction quality against state-of-the-art methods to highlight our method's practical advantages in both fidelity and efficiency.

### Effectiveness of Detail Preservation Network

To evaluate the contribution of the Detail Preservation Network (DPN), we analyze its impact on both anomaly detection performance and reconstruction quality. As shown in Table 2 of the main paper, incorporating DPN (without CRD) improves image-level and pixel-level AUROC from 94.1% / 95.8% to 95.7% / 96.6%.

For reconstruction quality evaluation, we compute PSNR and SSIM on normal images using the final recursive output ($n = 5$), with and without DPN. As summarized in Table S5, DPN yields average SSIM and PSNR gains of 0.059 and 0.61 dB, respectively, with maximum improvements reaching 0.32 and 2.61 dB. Additionally, when DPN is combined with CRD, it leads to further AUROC improvements of 0.5% (image-level) and 0.7% (pixel-level), demonstrating its contribution to anomaly detection performance.

Figure S1 further provides a qualitative comparison of reconstruction results with and without DPN on representative samples. The standard RcAE output $\mathbf{I}_R^5$ often suffers from loss of fine details, especially on highly textured surfaces such as *leather* and *zipper*. In contrast, the DPN-enhanced output $\mathbf{I}_D^5$ restores critical texture and structural information, while maintaining anomaly suppression. This selective detail enhancement capability arises from DPN's independent training on normal samples, which prevents rein-

Table S5: Quantitative evaluation of DPN. Inference cost is measured per image.

| Model | Inference Cost | Quality Gain | | AUROC Gain |
|---|---|---|---|---|
| | | Max. | Avg. | |
| +DPN | +2.1 ms | SSIM: +0.32<br>PSNR: +2.61 dB | SSIM: +0.059<br>PSNR: +0.61 dB | I: +0.5<br>P: +0.7 |

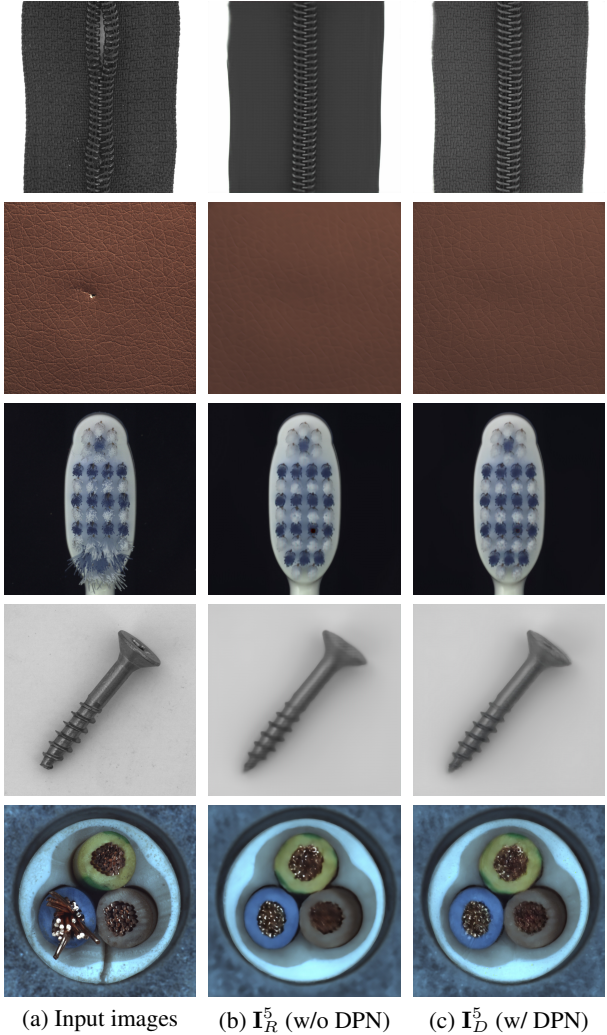

(a) Input images    (b) $\mathbf{I}_R^5$ (w/o DPN)    (c) $\mathbf{I}_D^5$ (w/ DPN)

Figure S1: Qualitative comparison of reconstruction results with/without DPN. DPN enhances fine details while preserving anomaly suppression, particularly on textured surfaces such as *leather* and *zipper*.

troduction of anomalies. However, while the improvements are most notable on highly textured surfaces, such as *leather* and *zipper*, the effect of DPN is less pronounced on objects with smooth or homogeneous textures (e.g., *bottle* and *grid*), where detail loss is minimal.

### Reconstruction Dynamics of the RcAE

As discussed in Section 3 in the main paper, the recursive convolutional autoencoder (RcAE) progressively transforms the input into an anomaly-free reconstruction via shared parameters across recursive iterations. A unique advantage of our framework is its ability to utilize intermediate reconstructions $\{\mathbf{I}_D^1, \ldots, \mathbf{I}_D^N\}$ for anomaly detection.

Figure S2 illustrates the reconstruction results at different recursion depths. Early iterations retain more local details but may preserve some anomalies, whereas later iterations suppress anomalies more effectively but may lose fine structures. This trade-off enables our method to detect both small and large anomalies by aggregating multi-scale residual information across iterations. This hierarchical processing offers superior semantic representation compared to conventional single-pass autoencoders.

### Comparison of Reconstruction Quality

Figure S3 presents a visual comparison of reconstruction results from our method and several state-of-the-art approaches on MVTec AD. For fair comparison with conventional single-pass methods, we use the final recursive output $\mathbf{I}_D^5$ from our framework.

Among non-diffusion baselines, DREAM, EdgRec, and UniAD exhibit limited capability in anomaly removal, often leaving visible defects in reconstructed images. Additionally, these methods tend to over-smooth textures, resulting in loss of fine structural details.

Diffusion-based methods such as DiAD and GLAD generally offer higher reconstruction fidelity. However, DiAD can effectively suppress many anomalies, but in some cases with huge structure anomalies (i.e., *transistor*) it fails to fully remove abnormal patterns, leaving subtle traces in the output. GLAD achieves strong reconstruction quality, but occasionally introduces semantic or details inconsistencies (such as *metal nut* and *grid*), potentially due to the stochastic nature of diffusion sampling. Moreover, both DiAD and GLAD incur substantial computational costs due to their iterative sampling steps, which often requiring hundreds of forward passes per image, causing in high latency and significant resource demands. This complexity limits their applicability in real-time or edge-computing scenarios.

In contrast, our method delivers high-fidelity reconstructions that are both detail-preserving and anomaly-free, closely matching the quality of GLAD. Notably, this is achieved with substantially lower computational cost and fewer parameters, enabling practical deployment in resource-constrained environments (see Figure 4 in the main paper). These results underscore the strength of our approach in balancing visual quality, detection accuracy, and inference efficiency.

Furthermore, as shown in Table 1 of the main paper, our method achieves the highest localization performance and competitive detection accuracy across all datasets, demonstrating strong and consistent generalization.

## Broader Impact Discussion

We proposed an efficient recursive autoencoder framework for unsupervised industrial anomaly detection, achieving state-of-the-art performance with a few computational cost and no external priors. Its lightweight, self-supervised design is ideal for real-world industrial applications such as automated defect inspection, predictive maintenance, and equipment monitoring, especially where computational resources or annotated data are limited.

Beyond industrial use cases, the core principles of our approach in this paper (i.e., recursive reconstruction refine-

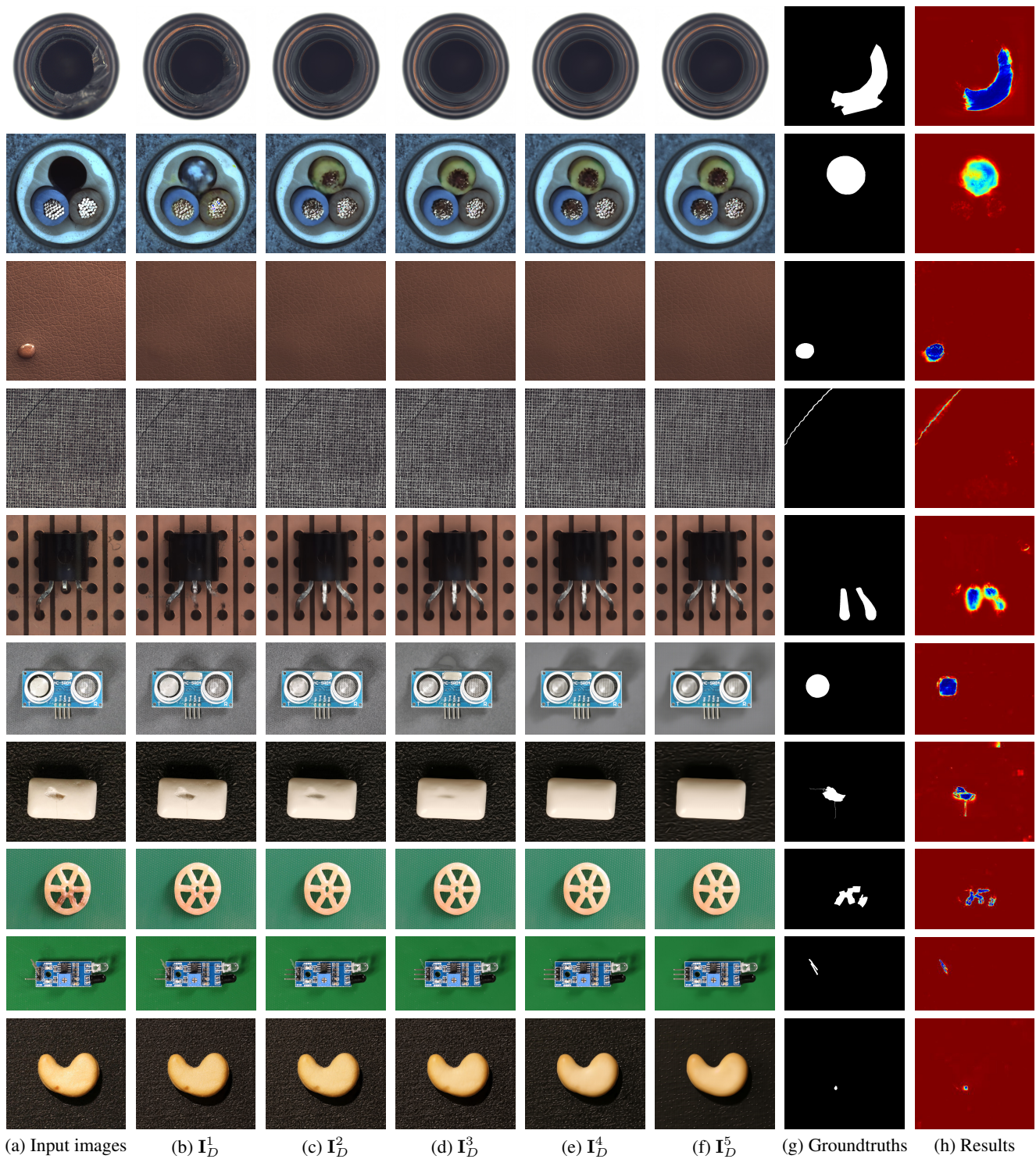

Figure S2: Qualitative reconstruction results at different recursion steps ($n = 1$ to $5$) using RcAE ($N = 5$), along with ground truth masks and final anomaly maps. Early iterations (b–d) retain more local details but may preserve anomalies. Later iterations (e–f) progressively normalize anomalies but can lose fine textures. The final anomaly map leverages multi-scale features across all recursive outputs $\{\mathbf{I}_D^1, \ldots, \mathbf{I}_D^5\}$, enabling robust detection of both small and large anomalies.

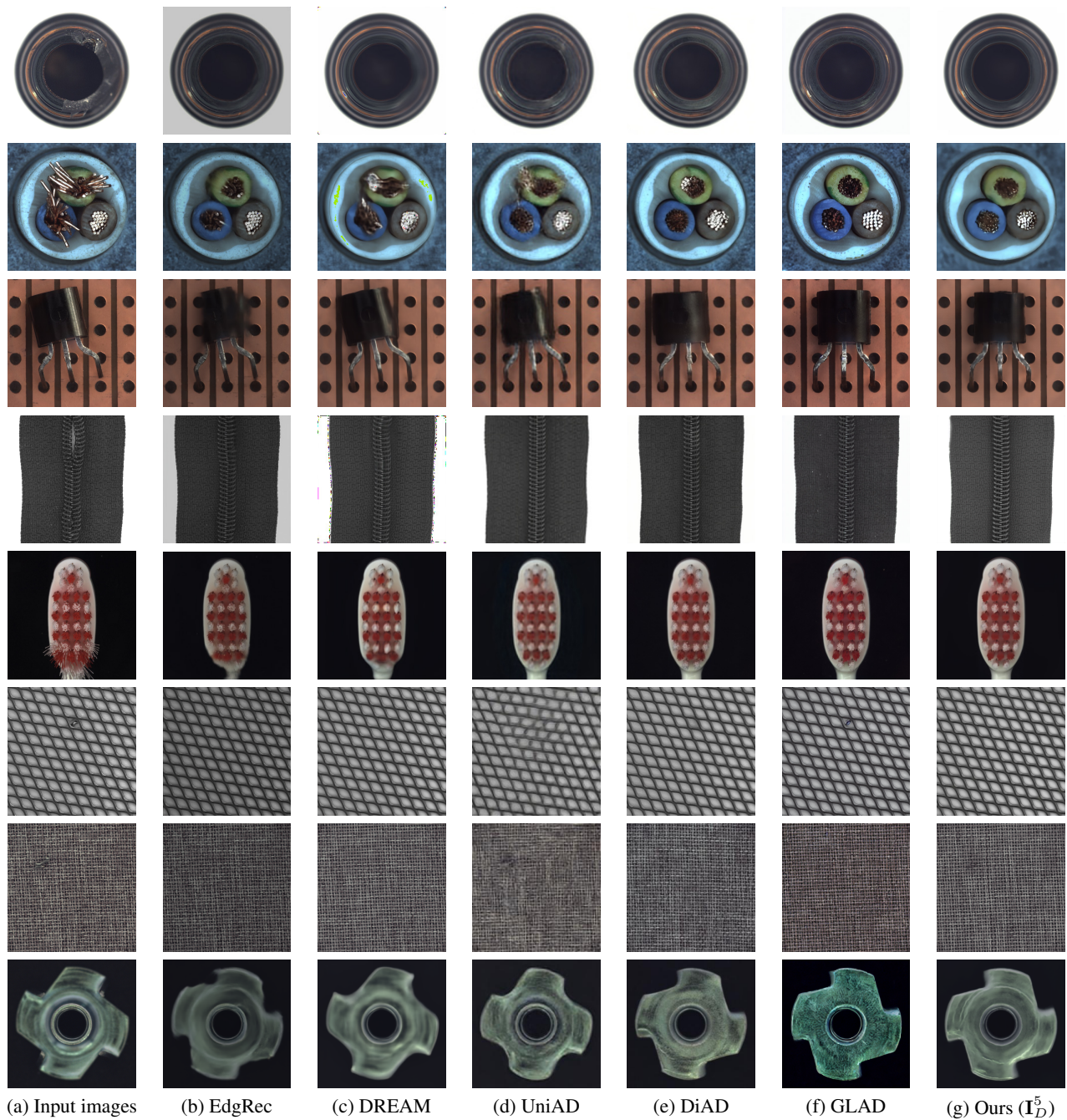

Figure S3: Comparison of reconstruction results between our method and prior work. For our method, the final recursive reconstruction $\mathbf{I}_D^5$ is used. Non-diffusion methods (b–d) fail to fully suppress anomalies and recover fine details. Diffusion-based methods (e–f) achieve high fidelity but at high computational cost, and GLAD occasionally introduces semantic or details inconsistencies. Our method (g) achieves comparable visual quality to GLAD while being significantly more efficient.

ment, parameter sharing, and multi-scale anomaly localization) offer broad applicability. Potential extensions include medical imaging (e.g., anomaly detection in radiology), remote sensing (e.g., disaster assessment), and AI security (e.g., defense of backdoor attack in vision models), where efficient, unsupervised solutions are in high demand. Our future work will explore integrating lightweight domain priors or hybrid architectures to enhance semantic understanding and extend our recursive framework to other vision tasks requiring both precision and efficiency.



\end{document}